**Federated learning model for predicting major postoperative complications**

Yonggi Park, PhD[a,b,‖], Yuanfang Ren, PhD[a,b,‖], Benjamin Shickel, PhD[a,b], Ziyuan Guan, MS[a,b], Ayush Patel[a], Yingbo Ma, PhD[a,b], Zhenhong Hu, PhD[a,b], Tyler J. Loftus, MD, PhD[a,c], Parisa Rashidi, PhD[a,d], Tezcan Ozrazgat-Baslanti, PhD[a,b,*], Azra Bihorac, MD, MS[a,b,*]

[‖]These first authors have contributed equally

[*]These senior authors have contributed equally

[a] University of Florida Intelligent Clinical Care Center, Gainesville, FL.

[b] Department of Medicine, Division of Nephrology, Hypertension, and Renal Transplantation, University of Florida, Gainesville, FL.

[c] Department of Surgery, University of Florida, Gainesville, FL.

[d] Department of Biomedical Engineering, University of Florida, Gainesville, FL.

Corresponding author: Azra Bihorac, MD, MS, University of Florida Intelligent Clinical Care Center, Division of Nephrology, Hypertension, and Renal Transplantation, Department of Medicine, University of Florida, PO Box 100224, Gainesville, FL 32610-0224. Telephone: (352) 294-8580; Fax: (352) 392-5465; Email: abihorac@ufl.edu

**Abstract**

**Background:** The postoperative complication is associated with a significant mortality rate in the United States annually. The accurate prediction of postoperative complication risk using Electronic Health Records (EHR) and artificial intelligence shows great potential. Training a robust artificial intelligence model typically requires large-scale and diverse datasets. In reality, collecting medical data often encounters challenges surrounding privacy protection.

**Methods:** This retrospective cohort study includes adult patients who were admitted to UFH Gainesville (GNV) (n = 79,850) and Jacksonville (JAX) (n = 28,636) for any type of inpatient surgical procedure. Using perioperative and intraoperative features, we developed federated learning models to predict nine major postoperative complications including prolonged (>48 hours) intensive care unit stay and mechanical ventilation, neurological complications including delirium, cardiovascular complications, acute kidney injury, venous thromboembolism, sepsis, wound complications that include infectious and mechanical wound complications, and hospital mortality. We compared federated learning models with local learning models trained on a single site and central learning models trained on pooled dataset from two centers.

**Results:** Our federated learning models achieved the area under the receiver operating characteristics curve (AUROC) values ranged from 0.81 for wound complications to 0.92 for prolonged ICU stay at UFH GNV center. At UFH JAX center, these values ranged from 0.73-0.74 for wound complications to 0.92-0.93 for hospital mortality. Federated learning models achieved comparable AUROC performance to central learning models, except for prolonged ICU stay, where the performance of federated learning models was slightly higher than central learning models at UFH GNV center, but slightly lower at UFH JAX center. In addition, our federated learning model obtained comparable performance to the best local learning model at each center, demonstrating strong generalizability.

**Conclusion:** Federated learning is shown to be a useful tool to train robust and generalizable models from large scale data across multiple institutions where data protection barriers are high.

**Introduction**

From an estimated 234.2 million major surgical procedures worldwide, an approximate 1 million patients die during or immediately after surgery every year.[1, 2] In the United States specifically, over 15 million major, inpatient surgeries are performed and at least 150,000 patients die within 30 days after surgery each year due to postoperative complications.[3] Postoperative complications occur up to 32% of surgeries.[4] Beyond the immediate health impacts on patients, these complications incur substantial financial costs, both for individuals and healthcare institutions. The accurate prediction of postoperative complication risk during the preoperative period is crucial, as it allows for the identification of patients who would benefit from post-surgery admission to an intensive care unit as well as more informed decision-making by patients and their healthcare providers regarding the appropriateness of surgery and treatment plan, consequently resulting in improved overall patient outcomes.

Assessing surgical risk necessitates the timely and accurate synthesis of vast clinical information across the perioperative care continuum. The integration of artificial intelligence (AI) into decision support tools has the ability to enhance the reliability, consistency, and accuracy of risk assessments. Furthermore, the widespread adoption of electronic health records (her) has facilitated data digitization, enabling the increased utilization of machine learning tools for risk surveillance and diagnosis.[5, 6] Although numerous studies have yielded significant insights into postoperative complication risk estimation using single-hospital data, there is a need for broader representation from diverse populations to enhance generalizability.[4, 6-8] This is crucial for applying machine learning in accurately predicting outcomes for target populations.[4, 6, 8] Even for those with the necessary resources, conventional approaches involving centralized data pooling from multiple centers raise concerns related to patient privacy and data protection.[9-12] Due to these concerns, federated learning is an optimal approach to effectively predict postoperative complications. This collaborative and decentralized machine learning approach involves maintaining and enhancing a deep-learning model within a central server, while the training process is distributed across various medical centers.[12, 13] Each center develops an individualized model based on its unique patient data, and the central server updates the global model following predefined criteria, selectively incorporating valuable feedback.[13] Importantly, this novel, scalable, and sustainable method facilitates collaborative training among multiple medical centers, enabling batch-wise training of machine learning models with distributed datasets across all institutions, while addressing concerns related to patient data privacy and upholding privacy and security standards.[8-13] Despite successful applications of federated learning in predicting outcomes such as acute kidney injury (AKI) stage, adverse drug reactions, hospitalizations, and COVID-19 mortality, its application for predicting surgical postoperative outcomes has not been explored.[8, 9, 12-14]

Our objective was to develop a robust, generalizable federated learning model to accurately predict the risk of postoperative complications using EHR data from two academic hospitals. Our hypothesis was that our federated learning model trained on distributed data would have comparable performance to a model trained on pooled data while preserving data privacy and security, and exhibit enhanced generalizability.

**Methods**

*Data Source and Participants*

Using the University of Florida Health (UFH) Integrated Repository as an honest broker for data de-identification, we gathered two single-center, longitudinal EHR datasets for all adult patients who were admitted to UFH Gainesville (GNV) and Jacksonville (JAX) for any type of inpatient surgical procedure between January 1, 2012, and May 1, 2021. We excluded minor procedures performed for controlling pain, gastrointestinal related minor surgeries and organ donation surgeries. Detailed inclusion and exclusion criteria used to identify encounters involving completed inpatient surgical procedures are shown in eFigure 1 in the Supplement. When a patient had multiple surgeries during one admission, only the first surgery was used in the analysis. The final retrospective UFH GNV cohort consisted of 62,827 patients undergoing 79,850 surgeries; the retrospective JAX GNV cohort consisted of 23,563 patients undergoing 28,636 surgeries.

Each dataset includes demographic information, vital sign, laboratory values, medications, diagnoses and procedures for all index admissions as well as admissions within 12 months before index admissions. This study was approved by the University of Florida Institutional Review Board and Privacy Office (IRB#201600223, IRB#201600262) as an exempt study with a waiver of informed consent.

*Study Design*

We followed the guidelines given in Transparent Reporting of a multivariable prediction model for Individual Prognosis or Diagnosis (TRIPOD)[15] and Leisman et al. under the type 2b analysis category[16]. We chronologically split each dataset by surgery start dates into three cohorts: training (63% of observations, n=51,953 surgeries for UFH GNV and n=18,502 surgeries for UFH JAX), validation (7% of observations, n=5,565 surgeries for UFH GNV and n=2,023 surgeries for UFH JAX) and test (30% of observations, n=22,332 surgeries for UFH GNV and n=8,111 surgeries for UFH JAX) cohorts to mitigate potential adverse effects of dataset drift due to changes in clinical practice or patient populations over time. We developed the model using training cohort and employed three distinct learning paradigms for training: local learning, central learning and federated learning (Figure 1). Local learning trains the model on isolated data of each center, while central learning trains the model on centralized data pooled from multiple centers. Federated learning trains each local model at each center and aggregates the respective computed model weights at a server without the need to exchange the actual data itself. We utilized validation cohort to tune the hyper-parameters and select the model. We validated the performance of models on test cohort.

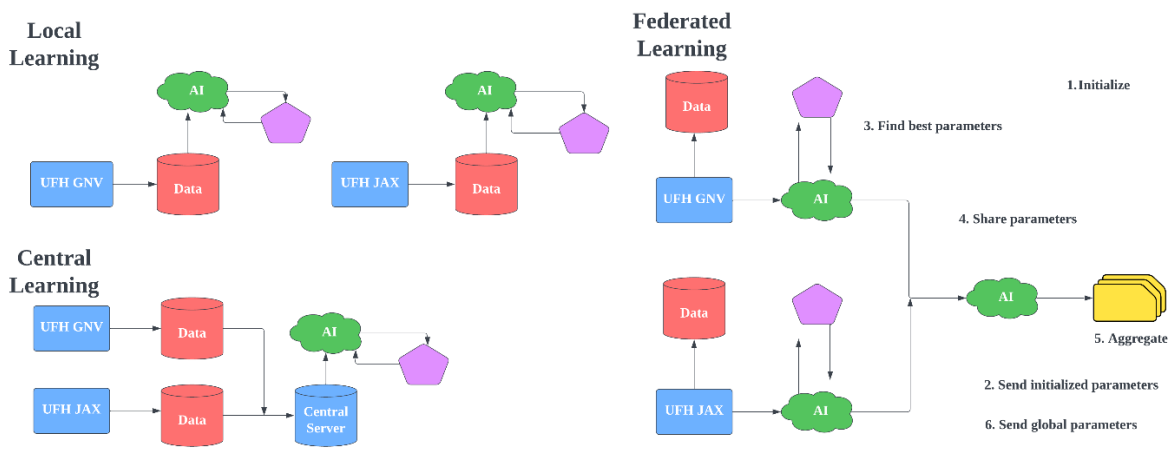

**Figure 1. Difference in learning paradigms of local learning, central learning and federated learning.** Local learning trains the model on isolated data of each center. Central learning trains the model on centralized data pooled from multiple centers. Federated learning trains each local model at each center and aggregates the respective computed model weights at a server without the need to exchange the actual data itself.

*Outcome*

We have developed and implemented an array of AI algorithms for computable phenotyping and dynamic perioperative risk assessment for nine postoperative complications including prolonged (>48 hours) intensive care unit (ICU) stay and mechanical ventilation (MV), neurological complications including delirium, cardiovascular complications, AKI, venous thromboembolism (VTE), sepsis, wound complications that include infectious and mechanical wound complications, and hospital mortality.[4, 7, 17] We used the exact dates to calculate the duration of MV and ICU stay. We defined AKI using KDIGO consensus criteria[18, 19] while a set of previously described criteria was applied to annotate the remaining complications. The algorithm also calculates risk probabilities for hospital mortality, where the date of death was determined using hospital records and the search of the Social Security Death Index.[20]

*Predictors*

For each patient we considered all potential predictors available in the harmonized dataset. We identified 402 preoperative features, including demographic, socio-economic, admission information (e.g., admission source), planned procedure and provider information, comorbidity, medications, and laboratory measurements. eTable 1 in the Supplement provides a comprehensive list of all preoperative features along with their distribution across two cohorts. We derived patient comorbidities and calculated the composite Charlson comorbidity index using up to 50 International Classification of Disease (ICD) codes from all available historical diagnosis codes.[21-23] We extracted medications dispensed in the one year prior to scheduled surgery day using RxNorm data categorized into drug classes based on the United States (US) Department of Veterans Affairs National Drug File Reference Terminology[24]. We extracted laboratory values measured one year prior to surgery using Logical Observation Identifiers Names and Codes (LOINC) and calculated derived features, including the count, mean, variance, minimum and maximum values. We also identified nine intraoperative time series data including systolic blood pressure, diastolic blood pressure, mean arterial pressure, heart rate, temperature, end-tidal carbon dioxide, peripheral capillary oxygen saturation, peak inspiratory pressure, respiratory rate, and minimum alveolar concentration. eFigure 2 presents the distribution of these intraoperative time series data across two centers. For continuous preoperative variables and intraoperative time series variables, we also created 'presence' features to enable our model to distinguish if a given value was observed or imputed, acknowledging that the patterns of missingness might be informative.

*Data Preprocessing*

We processed feature variables to remove outliers, impute missingness and standardize the values.[6, 25] Set of automatic rules was used for the removal of outliers that were considered unreasonable observations by medical experts. For continuous variables, observations in the top and bottom 1% of the distribution were considered as outliers, and they were imputed with a random number generated from the range between 0.5th and 5th percentiles and between 95th and 99.5th percentiles, respectively. All missing observations were imputed using automated algorithm. For nominal variables with missing entries, a distinct "missing" category was created. For continuous variables, the median value from the training cohort for a given variable was used for imputation. Intraoperative time series data was resampled at one-minute intervals. Missing time series data was processed by linear interpolation. If an entire time series variable was missing, the median values from training cohort was used for imputation. We standardized the continuous variables using standard normalization techniques. For local learning and federated learning, the entire data preprocessing process was independently carried out at each center. For central learning, the process was conducted at pooled data from two centers.

*Model Architecture*

Following the model architecture in the study[6], we developed two deep learning models to assess the risk of postoperative complications: a preoperative model and a postoperative model. The preoperative model utilized only preoperative features, whereas the postoperative model was developed using a combination of both preoperative and intraoperative features.

The postoperative model consisted of two sub-models: one for processing preoperative features and another for intraoperative features. In the preoperative sub-model, features were categorized into continuous, binary and high-cardinality type and then input into respective neural networks tailored for each variable type. The network architecture for continuous and binary features consisted of fully connected layers, while for high-cardinality features, it involved embedding layers followed by fully connected layers. The latent representation from three neural networks was combined through a fully connected layer. For the intraoperative sub-model, a bidirectional recurrent neural network equipped with gated recurrent units and an attention mechanism was employed to process intraoperative multivariate time series data. The final patient representation was derived by integrating the outputs from both sub-models through another fully connected layer. This integrated output was then passed into nine distinct branches, each corresponding to one of the nine outcomes, to calculate the risk probability. Within each branch, there was an outcome-specific fully connected layer followed by a sigmoid activation function, generating a score for each outcome. This score was interpreted as the likelihood of a preoperative patient developing a specific postoperative complication. For more detailed description of the model, we refer readers to Shickel et al.[6]

The preoperative model shared a similar architecture with postoperative model but differed in that it did not include the intraoperative sub-model. We trained these two models using three learning paradigms: local learning, central learning and federated learning. We used three federated learning algorithms including FedAvg[26], FedProx[27], and SCAFFOLD[28]. FedAvg aggregating model updates from multiple clients using a central server, is a popular federated learning approach for its simplicity, scalability and efficiency in communication. Despite these strengths, it encounters challenges pertaining to data heterogeneity and potential security vulnerabilities. FedProx extends the FedAvg algorithm by adding a regularization term to loss function, which encourages model similarity across clients, thereby addressing the challenges posed by data and model heterogeneity. SCAFFOLD addresses the problem of non-Independently and Identically Distributed (Non-IID) data in federated learning, but it comes with the trade-offs of higher computational and communication demands.

*Statistical Analysis*

We evaluated the robustness of models by 1) performing subgroup analysis based on sex (female vs male), race (African American vs non-African American) and age (age≤65 vs >65 years old); 2) conducting a sensitivity analysis of downsampling the data from the UFH GNV center to equalize the sample sizes between the two centers. We repeated the sampling experiment 10 times to enhance the reliability of the experiment's findings.

We assessed each model's discrimination using area under the receiver operating characteristic curve (AUROC) and area under the precision-recall curve (AUPRC). To obtain 95% confidence intervals for all performance metrics, we employed bootstrap sampling and non-parametric methods. For comparing clinical characteristics and outcomes of patients, we utilized the χ2 test for categorical variables and the Kruskal-Wallis test for continuous variables. The threshold for statistical significance was set at less than 0.05 for 2-sided tests. We adjusted p values for the family-wise error rate resulting from multiple comparisons using the Bonferroni correction. Data analysis was conducted using Python software of version 3.9, R software of version 4.3.3 and NVFlare of version 2.3.

**Results**

*Patient Baseline Characteristics and Outcomes*

Among 39,582 adult patients who received 51,953 major inpatient surgical procedures in the UFH training cohort, the mean (standard deviation, SD) age was 57 (17) years; 19,698 patients (50%) were female, and 19,884 (50%) were male (Table 1); 30,947 patients (78%) were White and 5,474 (14%) were African American. The UFH JAX training cohort contained 18,502 inpatient surgical procedures involving 14,846 patients and the demographic characteristics was significantly different from the UFH GNV training cohort (younger patients with mean (SD) age, 52 (17) years; 7,204 (49%) were females and 7,642 (51%) were males; 8,675 (59%) were White and 5,158 (35%) were African American).

Each major type of procedure, such as cardiothoracic, gastrointestinal, neurological, obstetric, oncological, otolaryngological, urological, and vascular surgeries, was significantly represented (eTable 1 in the Supplement). The prevalence of postoperative complications in the UFH GNV training cohort was 15% for AKI, 12% for cardiovascular complications, 17% for neurological complications including delirium, 29% for prolonged ICU stay, 9% for prolonged

12MV, 8% for sepsis, 5% for VTE, 16% for wound complications, and 2% for hospital mortality. Compared with UFH GNV training cohort, UFH JAX training cohort had similar complication prevalence, except neurological complications including delirium (12% vs. 17%), prolonged ICU stay (24% vs. 29%), and wound complications (14% vs. 16%). There was also some variation in complication prevalence between the training and test cohorts at each center (eTable 2 in the Supplement). For example, the prevalence of all complications except hospital mortality in UFH GNV training cohort was significantly different from that in UFH GNV test cohort. At UFH JAX center, the prevalence of complications including cardiovascular complications, prolonged mechanical ventilation, venous thromboembolism and hospital mortality was significantly different between training and test cohorts.

Distribution of features used for model development was significantly different between two centers (eTable 1 and eFigure 2 in the Supplement). Additional details of patient clinical characteristics, complication prevalence and distribution of features were shown in Table 1 and eTables 1 and 2 in the Supplement.

**Table 1. Patient Characteristics in Training Cohorts**

| Variables | UFH GNV | UFH JAX | p-value |
|---|---|---|---|
| Number of patients, n | 39,582 | 14,846 | |
| Number of surgical procedures, n | 51,953 | 18,502 | |
| Age in years, mean (SD)[a] | 57 (17) | 52 (17) | <.001 |
| Sex, n (%)[a] | | | |
|   Male | 19,884 (50) | 7,642 (51) | <.001 |
|   Female | 19,698 (50) | 7,204 (49) | <.001 |
| Race, n (%)[a,b] | | | |
|   White | 30,947 (78) | 8,675 (59) | <.001 |
|   African American | 5,474 (14) | 5,158 (35) | <.001 |
|   Other[c] | 2,521 (6) | 952 (6) | 0.32 |
|   Missing | 640 (2) | 61 (0) | <.001 |
| Ethnicity, n (%)[a,b] | | | |
|   Non-Hispanic | 37,165 (94) | 14,079 (95) | <.001 |
|   Hispanic | 1,707 (4) | 676 (5) | 0.96 |
|   Missing | 710 (2) | 91 (0) | <.001 |
| Marital Status, n (%)[a] | | | |
|   Married | 19,182 (48) | 5,055 (34) | <.001 |
|   Single | 11,769 (30) | 5,392 (36) | <.001 |
|   Divorced | 6,084 (15) | 3,875 (26) | <.001 |
|   Missing | 2,547 (7) | 524 (4) | <.001 |
| Insurance, n (%)[a] | | | |
|   Medicare | 17,767 (45) | 4,408 (30) | <.001 |
|   Private | 12,311 (31) | 4,308 (29) | <.001 |
|   Medicaid | 6,296 (16) | 5,908 (40) | <.001 |
|   Uninsured | 3,208 (8) | 222 (1) | <.001 |
| Complications, n (%)[d] | | | |
|   Acute kidney injury | 7,924 (15) | 2,577 (14) | <.001 |
|   Cardiovascular complications | 6,419 (12) | 2,104 (11) | <.001 |
|   Neurological complications, including delirium | 8,873 (17) | 2,199 (12) | <.001 |
|   Prolonged ICU stay | 15,049 (29) | 4,372 (24) | <.001 |
|   Prolonged mechanical ventilation | 4,667 (9) | 1,492 (8) | <.001 |
|   Sepsis | 3,958 (8) | 1,601 (9) | <.001 |
|   Venous thromboembolism | 2,483 (5) | 721 (4) | <.001 |
|   Wound complications | 8,088 (16) | 2,594 (14) | <.001 |
|   Hospital mortality | 952 (2) | 329 (2) | 0.66 |

Abbreviation: ICU, intensive care unit.
[a] Data were reported based on values calculated at the latest hospital admission.
[b] Race and ethnicity were self-reported.
[c] Other races include American Indian or Alaska Native, Asian, Native Hawaiian or Pacific Islander, and multiracial.
[d] Data were reported based on postoperative complication status for each surgical procedure.

*Comparison between Central Learning and Federated Learning Models*

We evaluated the performance of central learning models and federated learning models by calculating AUROC and AUPRC values (Table 2 and eTable 3 in the supplement). In the preoperative federated learning models, AUROC values ranged from 0.79-0.80 for wound complications to 0.89-0.90 for hospital mortality at UFH GNV center. At UFH JAX center, these values ranged from 0.71 for wound complications to 0.90 for hospital mortality. In the postoperative federated learning models, AUROC values ranged from 0.81 for wound complications to 0.92 for prolonged ICU stay at UFH GNV center. At UFH JAX center, these values ranged from 0.73-0.74 for wound complications to 0.92-0.93 for hospital mortality. All postoperative models utilizing intraoperative time series vitals outperformed preoperative models in both sites in terms of AUROC, especially for cardiovascular complications (UFH GNV: 0.85 vs 0.82; UFH JAX: 0.84 vs 0.78-0.79). Federated learning models achieved comparable AUROC performance to central learning models, except for prolonged ICU stay, where the performance of federated learning models was slightly higher than central learning models (preoperative models: 0.90 [95% CI, 0.90-0.91] vs 0.89 [95% CI, 0.89-0.90]; postoperative models: 0.92 [95% CI, 0.92-0.92] vs 0.91 [95% CI, 0.91-0.92]) at UFH GNV center, but slightly lower at UFH JAX center (preoperative models: 0.87 [95% CI, 0.86-0.88] vs 0.89 [95% CI, 0.89-0.90]; postoperative models: 0.89 [95% CI, 0.88-0.90] vs 0.91 [95% CI, 0.90-0.91]). Federated learning models achieved comparable AUPRC performance to central learning model at UF GNV center, while yielding lower discrimination at UFH JAX center for the majority of the complications (eTable 3 in the Supplement). Three federated learning models had similar performance and SCAFFOLD model achieved the overall best performance. For simplicity, we only reported the performance of SCAFFOLD model in the following sections.

**Table 2. Comparison of AUROC with 95% confidence interval for central learning and federated learning models**

|  |  | UFH GNV | | | | UFH JAX | | | |
|---|---|---|---|---|---|---|---|---|---|
| Outcome | Period | CL | FedAvg | FedProx | SCAFFOLD | CL | FedAvg | FedProx | SCAFFOLD |
| Prolonged ICU stay | PreOp | 0.89 (0.89-0.90) | 0.90 (0.90-0.90)[a] | 0.90 (0.90-0.91)[a] | 0.90 (0.90-0.91)[a] | 0.89 (0.89-0.90) | 0.87 (0.87-0.88)[a] | 0.87 (0.86-0.87)[a] | 0.87 (0.86-0.87)[a] |
|  | PostOp | 0.91 (0.91-0.92) | 0.92 (0.92-0.92)[a] | 0.92 (0.92-0.92)[a] | 0.92 (0.92-0.92)[a] | 0.91 (0.90-0.91) | 0.89 (0.88-0.89)[a] | 0.89 (0.88-0.90)[a] | 0.90 (0.89-0.90)[a] |
| Sepsis | PreOp | 0.88 (0.87-0.88) | 0.88 (0.87-0.88) | 0.88 (0.87-0.89) | 0.88 (0.87-0.88) | 0.89 (0.88-0.90) | 0.88 (0.87-0.89) | 0.88 (0.87-0.89) | 0.88 (0.87-0.89) |
|  | PostOp | 0.89 (0.88-0.89) | 0.89 (0.88-0.89) | 0.89 (0.88-0.89) | 0.89 (0.88-0.89) | 0.90 (0.89-0.91) | 0.89 (0.88-0.90) | 0.89 (0.88-0.90) | 0.89 (0.88-0.90) |
| Cardiovascular complication | PreOp | 0.82 (0.81-0.82) | 0.82 (0.81-0.82) | 0.82 (0.81-0.83) | 0.82 (0.81-0.82) | 0.80 (0.78-0.82) | 0.79 (0.77-0.80) | 0.78 (0.77-0.80) | 0.79 (0.78-0.80) |
|  | PostOp | 0.85 (0.84-0.85) | 0.85 (0.85-0.86) | 0.85 (0.85-0.86) | 0.85 (0.84-0.86) | 0.84 (0.83-0.85) | 0.84 (0.83-0.85) | 0.84 (0.82-0.85) | 0.84 (0.83-0.85) |
| Venous thromboembolism | PreOp | 0.83 (0.82-0.84) | 0.82 (0.81-0.83) | 0.83 (0.82-0.84) | 0.83 (0.82-0.84) | 0.81 (0.79-0.83) | 0.80 (0.78-0.82) | 0.79 (0.77-0.82) | 0.79 (0.77-0.81) |
|  | PostOp | 0.83 (0.82-0.84) | 0.84 (0.83-0.85) | 0.84 (0.83-0.85) | 0.84 (0.83-0.85) | 0.83 (0.81-0.85) | 0.82 (0.80-0.84) | 0.82 (0.80-0.84) | 0.83 (0.81-0.85) |
| Prolonged mechanical ventilation | PreOp | 0.90 (0.89-0.91) | 0.90 (0.89-0.91) | 0.90 (0.90-0.91) | 0.90 (0.90-0.91) | 0.84 (0.82-0.86) | 0.83 (0.82-0.85) | 0.82 (0.81-0.84) | 0.84 (0.82-0.85) |
|  | PostOp | 0.91 (0.91-0.92) | 0.92 (0.92-0.93) | 0.92 (0.91-0.93) | 0.92 (0.91-0.92) | 0.86 (0.85-0.88) | 0.85 (0.84-0.87) | 0.86 (0.84-0.87) | 0.86 (0.85-0.88) |
| Neurological complications, including delirium | PreOp | 0.85 (0.85-0.86) | 0.85 (0.85-0.86) | 0.85 (0.85-0.86) | 0.85 (0.85-0.86) | 0.84 (0.83-0.85) | 0.84 (0.83-0.85) | 0.83 (0.82-0.84) | 0.84 (0.83-0.85) |
|  | PostOp | 0.85 (0.85-0.86) | 0.86 (0.85-0.86) | 0.86 (0.85-0.86) | 0.86 (0.85-0.86)[a] | 0.85 (0.84-0.86) | 0.84 (0.83-0.85) | 0.84 (0.83-0.85) | 0.84 (0.83-0.85) |
| Wound complications | PreOp | 0.79 (0.78-0.80) | 0.79 (0.78-0.80) | 0.80 (0.79-0.80) | 0.80 (0.79-0.80) | 0.72 (0.70-0.73) | 0.71 (0.70-0.72) | 0.71 (0.70-0.73) | 0.71 (0.70-0.72) |
|  | PostOp | 0.80 (0.79-0.80) | 0.81 (0.80-0.81) | 0.81 (0.80-0.81)[a] | 0.81 (0.80-0.81) | 0.74 (0.72-0.75) | 0.73 (0.72-0.75) | 0.73 (0.72-0.75) | 0.74 (0.72-0.75) |
| Acute kidney injury | PreOp | 0.82 (0.81-0.82) | 0.82 (0.81-0.82) | 0.82 (0.82-0.83) | 0.82 (0.82-0.83) | 0.79 (0.78-0.81) | 0.79 (0.78-0.80) | 0.79 (0.78-0.80) | 0.79 (0.77-0.80) |
|  | PostOp | 0.82 (0.82-0.83) | 0.83 (0.82-0.84) | 0.83 (0.82-0.84) | 0.83 (0.83-0.84)[a] | 0.81 (0.80-0.82) | 0.81 (0.79-0.82) | 0.80 (0.79-0.82) | 0.81 (0.80-0.82) |
| Hospital mortality | PreOp | 0.90 (0.88-0.91) | 0.89 (0.87-0.90) | 0.90 (0.89-0.91) | 0.89 (0.87-0.90) | 0.92 (0.90-0.94) | 0.90 (0.87-0.92) | 0.90 (0.87-0.92) | 0.90 (0.88-0.92) |
|  | PostOp | 0.90 (0.89-0.91) | 0.91 (0.90-0.92) | 0.91 (0.90-0.92) | 0.91 (0.89-0.92) | 0.93 (0.91-0.95) | 0.92 (0.90-0.94) | 0.93 (0.91-0.95) | 0.92 (0.90-0.94) |

Abbreviation: ICU, intensive unit; CL, central learning; PreOp, preoperative; PostOp, postoperative.
[a] The p-values represent difference < 0.05 compared to central learning model for each site.

*Generalizability of Federated Learning Models*

We assessed the generalizability of our federated learning model by comparing its performance with two local learning models: the GNV model and the JAX model (Figure 2, Table 3 and eTable 4 in the Supplement). Each local learning model demonstrated strong performance within its respective center, but this performance declined when applied to a different center, indicating limited generalizability. That is, at UFH GNV center, GNV model had better AUROC performance than JAX model on all complications; while at UFH JAX center, compared to GNV model, JAX model achieved better AUROC performance on all complications except for wound complications (preoperative models: 0.68 [95% CI, 0.67-0.70] vs 0.70 [95% CI, 0.68-0.71]; postoperative models: 0.71 [95% CI, 0.70-0.72] vs 0.71 [95% CI, 0.70-0.72]).

On the other hand, our federated learning model obtained comparable performance to the best local learning model at each center, demonstrating strong generalizability. That is, at UFH GNV center, federated learning model achieved performance comparable to or slightly better than the best local learning model, GNV model, across all complications. Similar comparative results were observed for all complications at UFH JAX center, except for the prolonged ICU stay (preoperative models: 0.87 [95% CI, 0.86-0.87] vs 0.89 [95% CI, 0.89-0.90]; postoperative models: 0.90 [95% CI, 0.89-0.90] vs 0.91 [95% CI, 0.91-0.92]).

*Robustness of Federated Learning Models*

To determine if the performance of federated learning model was independent of patient characteristics, we conducted subgroup analysis based on sex, race and age (eTables 5-8). The performance of federated learning model showed significant differences between female and male patients in predicting sepsis, cardiovascular complication and VTE at both centers,

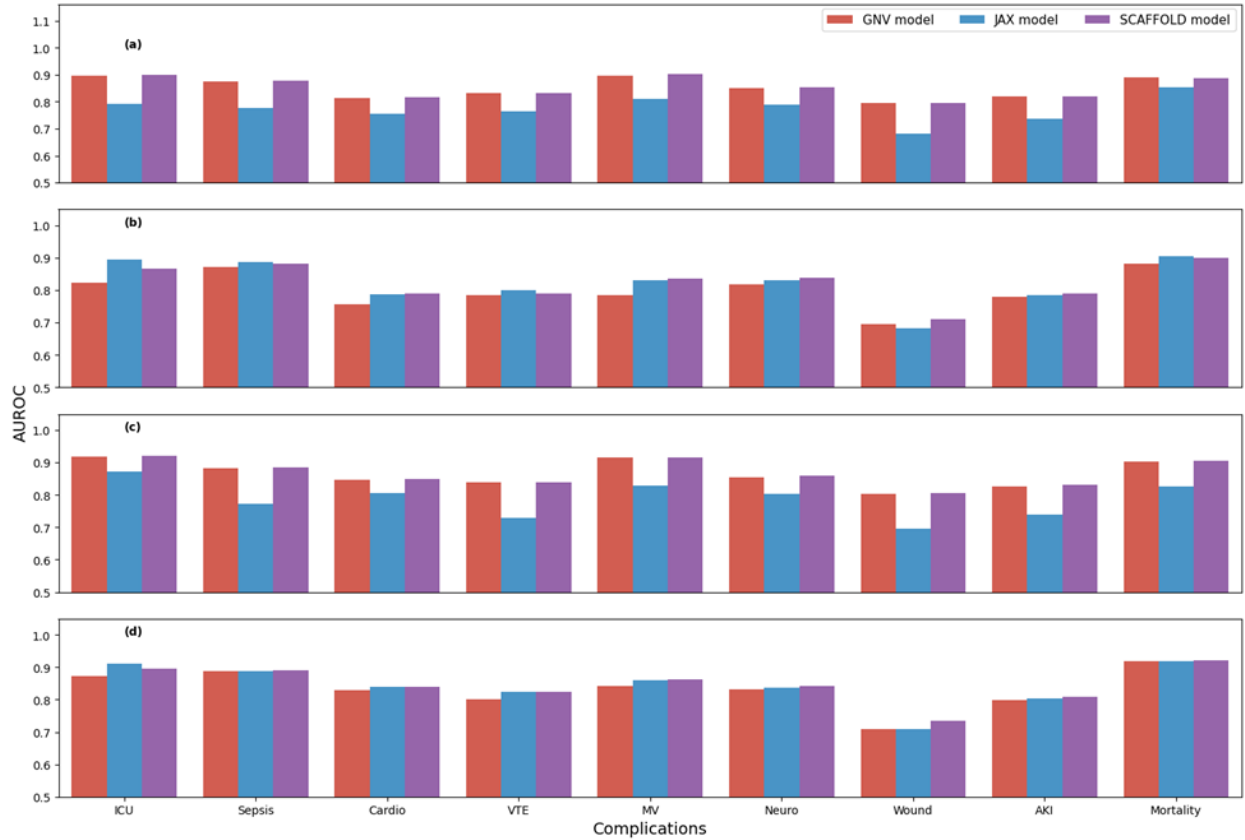

**Figure 2. Performance of local learning models and federated learning models.**
Preoperative models were evaluated on UFH GNV test data (a) and UFH JAX test data (b). Postoperative models were evaluated on UFH GNV test data (c) and UFH JAX test data (d). The full names of complications displayed on x-axis from left to right are: prolonged intensive care unit stay, sepsis, cardiovascular complications, venous thromboembolism, prolonged mechanical ventilation, neurological complications including delirium, wound complications, acute kidney injury, and hospital mortality.

**Table 3. Comparison of AUROC (95% CI) for local learning and federated learning models.**

| | | Preoperative models | | Postoperative models | |
|---|---|---|---|---|---|
| **Outcome** | **Model** | **GNV test data** | **JAX test data** | **GNV test data** | **JAX test data** |
| Prolonged ICU stay | GNV Model | 0.90 (0.89-0.90) | 0.82 (0.81-0.83)[a] | 0.92 (0.92-0.92) | 0.87 (0.87-0.88)[a] |
| | JAX Model | 0.79 (0.79-0.80)[a] | 0.89 (0.89-0.90)[a] | 0.87 (0.87-0.88)[a] | 0.91 (0.91-0.92)[a] |
| | SCAFFOLD | 0.90 (0.90-0.91) | 0.87 (0.86-0.87) | 0.92 (0.92-0.92) | 0.90 (0.89-0.90) |
| Sepsis | GNV Model | 0.87 (0.87-0.88) | 0.87 (0.86-0.89) | 0.88 (0.88-0.89) | 0.89 (0.88-0.90) |
| | JAX Model | 0.78 (0.77-0.79)[a] | 0.89 (0.88-0.90) | 0.77 (0.76-0.78)[a] | 0.89 (0.88-0.90) |
| | SCAFFOLD | 0.88 (0.87-0.88) | 0.88 (0.87-0.89) | 0.89 (0.88-0.89) | 0.89 (0.88-0.90) |
| Cardiovascular complication | GNV Model | 0.81 (0.81-0.82) | 0.76 (0.74-0.77)[a] | 0.85 (0.84-0.85) | 0.83 (0.82-0.84) |
| | JAX Model | 0.75 (0.75-0.76)[a] | 0.79 (0.77-0.80) | 0.81 (0.80-0.81)[a] | 0.84 (0.83-0.85) |
| | SCAFFOLD | 0.82 (0.81-0.82) | 0.79 (0.78-0.80) | 0.85 (0.84-0.86) | 0.84 (0.83-0.85) |
| Venous thromboembolism | GNV Model | 0.83 (0.82-0.84) | 0.79 (0.76-0.81) | 0.84 (0.83-0.85) | 0.80 (0.78-0.82) |
| | JAX Model | 0.77 (0.76-0.78)[a] | 0.80 (0.78-0.82) | 0.73 (0.71-0.74)[a] | 0.82 (0.80-0.84) |
| | SCAFFOLD | 0.83 (0.82-0.84) | 0.79 (0.77-0.81) | 0.84 (0.83-0.85) | 0.83 (0.81-0.85) |
| Prolonged mechanical ventilation | GNV Model | 0.90 (0.89-0.91) | 0.79 (0.77-0.80)[a] | 0.92 (0.91-0.92) | 0.84 (0.83-0.86) |
| | JAX Model | 0.81 (0.80-0.82)[a] | 0.83 (0.81-0.85) | 0.83 (0.82-0.84)[a] | 0.86 (0.84-0.88) |
| | SCAFFOLD | 0.90 (0.90-0.91) | 0.84 (0.82-0.85) | 0.92 (0.91-0.92) | 0.86 (0.85-0.88) |
| Neurological complications, including delirium | GNV Model | 0.85 (0.84-0.85) | 0.82 (0.81-0.83)[a] | 0.85 (0.85-0.86) | 0.83 (0.82-0.84) |
| | JAX Model | 0.79 (0.78-0.79)[a] | 0.83 (0.82-0.84) | 0.80 (0.80-0.81)[a] | 0.84 (0.83-0.85) |
| | SCAFFOLD | 0.85 (0.85-0.86) | 0.84 (0.83-0.85) | 0.86 (0.86-0.86) | 0.84 (0.83-0.85) |
| Wound complications | GNV Model | 0.80 (0.79-0.80) | 0.70 (0.68-0.71) | 0.80 (0.80-0.81) | 0.71 (0.70-0.72)[a] |
| | JAX Model | 0.68 (0.68-0.69)[a] | 0.68 (0.67-0.70)[a] | 0.70 (0.69-0.70)[a] | 0.71 (0.70-0.72)[a] |
| | SCAFFOLD | 0.80 (0.79-0.80) | 0.71 (0.70-0.72) | 0.81 (0.80-0.81) | 0.74 (0.72-0.75) |
| Acute kidney injury | GNV Model | 0.82 (0.81-0.83) | 0.78 (0.77-0.79) | 0.83 (0.82-0.83) | 0.80 (0.79-0.81) |
| | JAX Model | 0.74 (0.73-0.75)[a] | 0.78 (0.77-0.80) | 0.74 (0.73-0.75)[a] | 0.80 (0.79-0.82) |
| | SCAFFOLD | 0.82 (0.82-0.83) | 0.79 (0.78-0.80) | 0.83 (0.83-0.84) | 0.81 (0.80-0.82) |
| Hospital mortality | GNV Model | 0.89 (0.88-0.90) | 0.88 (0.86-0.91) | 0.90 (0.89-0.91) | 0.92 (0.90-0.94) |
| | JAX Model | 0.85 (0.84-0.87)[a] | 0.91 (0.88-0.93) | 0.83 (0.80-0.85)[a] | 0.92 (0.90-0.94) |
| | SCAFFOLD | 0.89 (0.87-0.90) | 0.90 (0.88-0.92) | 0.91 (0.89-0.92) | 0.92 (0.90-0.94) |

Abbreviation: ICU, intensive unit.
[a] The p-values represent difference < 0.05 compared to SCAFFOLD model for each site.

with a notable difference in hospital mortality prediction specifically at UFH JAX center. The performance was not affected by race across all complications at both centers, except for a significant difference observed in predicting sepsis at UFH GNV and hospital mortality at both centers. Age had a substantial impact on the model's performance, with this effect being more pronounced at the UFH GNV center.

To determine if the performance of federated learning model would be influenced by the various sample sizes across centers, we conducted a sensitivity analysis of downsampling the samples from UFH GNV centers to equalize the sample sizes between the two centers (eTable 8). We observed a pronounced performance decline at the larger data provider (UFH GNV), and a slight performance increase at UFH JAX center (prolonged ICU stay: 0.87 vs 0.89 $\pm$ 0.003).

## **Discussion**

We have developed preoperative and postoperative federated learning models to predict risks of major postoperative complications using large EHR datasets from two centers. The federated learning models proved to be more robust and generalizable than any local learning models. This improvement is attributed to the use of larger and more diverse datasets, which are crucial for developing robust and generalizable models. This underscores the benefits and effectiveness of federated learning paradigm. The federated learning models also proved to be not inferior to the central learning models while also maintaining data privacy and security.

Several studies have shown the utility and effectiveness of federated learning models in healthcare. There has been a significant number of studies exploring the applications of federated learning in COVID-19. Dayan et al. have proposed a federated learning model to predict the future oxygen requirements for patients with COVID-19 using vital signs, laboratory data and chest X-rays from 20 institutes.[29] The proposed model achieved an increase in generalizability of 38% when compared with local training models. Vaid et al. have presented a federated learning model to predict mortality in hospitalized patients with COVID-19 within 7 days.[8] Feki et al. have applied federated learning for COVID-19 screening using chest X-rays.[30] Federated learning has also been applied to enhance surgical outcomes. Feng at al. have developed a robust federated learning model for identifying high-risk patients with postoperative gastric cancer recurrence using computed tomography images.[31]

Several studies have shown that centers with small sample size can benefit from federated learning training from a larger and more diverse dataset.[29, 31] However, this trend wasn't mirrored at the UFH JAX center, despite its smaller sample count. A possible explanation is that the data size of UFH JAX cohort was sufficient to learn their own inherent data representation. In federated learning, updates from various clients are often weighted according to their sample sizes. We increased the weight of UFH JAX cohort in our sensitivity analysis, and observed limited improvements at UFH JAX center but obvious performance degradation for large data provider. We conjectured that balancing the data sizes across multiple centers to mitigate bias towards centers with smaller datasets is not advisable, and instead, fine-tuning trained federated learning model at each local center may be a better option.

The study had several limitations. First, it used data from only two centers, which offered limited diversity in populations and consequently restricted the generalizability of our findings. Second, the study was conducted as a simulation of federated learning, rather than through actual implementation. Certain practical challenges and dynamics of real-world deployment were not tested. For example, EHR data cleaning, labeling and standardization approaches may be various across centers. Other technical issues like network latency, data synchronization and varying computation capacities of nodes may significantly limit the possibility of federated learning in implementation. Future studies involving real implementation across multiple and diverse real-world settings is necessary to develop generalizable and robust models and validate the feasibility and efficiency of federated learning models in practical healthcare applications.

## **Conclusions**

In this work, we developed federated learning models for predicting major postoperative complications after surgery using EHR data from two centers. Our models achieved comparable performance to central learning models while maintaining data privacy and exhibited better robustness and generalizability than local learning models. In the future work, we aim to develop real implemented federated learning models using larger and more diverse datasets across multiple centers.

**Acknowledgments:** We gratefully acknowledge the technical support of NVIDIA AI Technology Center (NVAITC) at UF for this research. We would like to acknowledge the Intelligent Clinical Care Center research group for support provided for this study. We acknowledge the University of Florida Integrated Data Repository (IDR) and the UF Health Office of the Chief Data Officer for providing the analytic data set for this project.

**Funding:** A.B, P.R, T.O.B., B.S., and Y.R. were supported R01 GM110240 from the National Institute of General Medical Sciences (NIH/NIGMS).  This work was also supported in part by the NIH/NCATS Clinical and Translational Sciences Award to the University of Florida UL1 TR000064.  The content is solely the responsibility of the authors and does not necessarily represent the official views of the National Institutes of Health. The funders had no role in design and conduct of the study; collection, management, analysis, and interpretation of the data; preparation, review, or approval of the manuscript; and decision to submit the manuscript for publication. The authors declare that they have no conflict of interests. A.B. and T.O.B. had full access to all the data in the study and take responsibility for the integrity of the data and the accuracy of the data analysis.

References

1.	Lee PHU, Gawande AA. The number of surgical procedures in an American lifetime in 3 states. *Journal of the American College of Surgeons*. 2008;207(3):S75. doi:10.1016/j.jamcollsurg.2008.06.186
2.	Weiser TG, Regenbogen SE, Thompson KD, et al. An estimation of the global volume of surgery: a modelling strategy based on available data. *Lancet*. Jul 12 2008;372(9633):139-144. doi:10.1016/S0140-6736(08)60878-8
3.	Gawande AA, Regenbogen SE. Critical need for objective assessment of postsurgical patients. *Anesthesiology*. Jun 2011;114(6):1269-70. doi:10.1097/ALN.0b013e318219d76b
4.	Bihorac A, Ozrazgat-Baslanti T, Ebadi A, et al. MySurgeryRisk: Development and Validation of a Machine-learning Risk Algorithm for Major Complications and Death After Surgery. *Ann Surg*. Apr 2019;269(4):652-662. doi:10.1097/SLA.0000000000002706
5.	Juhn Y, Liu H. Artificial intelligence approaches using natural language processing to advance EHR-based clinical research. *J Allergy Clin Immunol*. Feb 2020;145(2):463-469. doi:10.1016/j.jaci.2019.12.897
6.	Shickel B, Loftus TJ, Ruppert M, et al. Dynamic predictions of postoperative complications from explainable, uncertainty-aware, and multi-task deep neural networks. *Sci Rep*. Jan 21 2023;13(1):1224. doi:10.1038/s41598-023-27418-5
7.	Datta S, Loftus TJ, Ruppert MM, et al. Added Value of Intraoperative Data for Predicting Postoperative Complications: The MySurgeryRisk PostOp Extension. *J Surg Res*. Oct 2020;254:350-363. doi:10.1016/j.jss.2020.05.007
8.	Vaid A, Jaladanki SK, Xu J, et al. Federated Learning of Electronic Health Records to Improve Mortality Prediction in Hospitalized Patients With COVID-19: Machine Learning Approach. *JMIR Med Inform*. Jan 27 2021;9(1):e24207. doi:10.2196/24207
9.	Choudhury O, Park Y, Salonidis T, Gkoulalas-Divanis A, Sylla I, Das AK. Predicting Adverse Drug Reactions on Distributed Health Data using Federated Learning. *AMIA Annu Symp Proc*. 2019;2019:313-322.
10.	Rieke N, Hancox J, Li W, et al. The future of digital health with federated learning. *NPJ Digit Med*. 2020;3:119. doi:10.1038/s41746-020-00323-1
11.	Yang Q, Liu Y, Chen T, Tong Y. Federated machine learning: Concept and applications. *ACM Transactions on Intelligent Systems and Technology (TIST)*. 2019;10(2):1-19.
12.	Brisimi TS, Chen R, Mela T, Olshevsky A, Paschalidis IC, Shi W. Federated learning of predictive models from federated Electronic Health Records. *Int J Med Inform*. Apr 2018;112:59-67. doi:10.1016/j.ijmedinf.2018.01.007
13.	Ng D, Lan X, Yao MM, Chan WP, Feng M. Federated learning: a collaborative effort to achieve better medical imaging models for individual sites that have small labelled datasets. *Quant Imaging Med Surg*. Feb 2021;11(2):852-857. doi:10.21037/qims-20-595
14.	Huang CT, Wang TJ, Kuo LK, et al. Federated machine learning for predicting acute kidney injury in critically ill patients: a multicenter study in Taiwan. *Health Inf Sci Syst*. Dec 2023;11(1):48. doi:10.1007/s13755-023-00248-5
15.	Collins GS, Reitsma JB, Altman DG, Moons KG. Transparent reporting of a multivariable prediction model for individual prognosis or diagnosis (TRIPOD): the TRIPOD Statement. *BMC Med*. Jan 6 2015;13:1. doi:10.1186/s12916-014-0241-z
16.	Leisman DE, Harhay MO, Lederer DJ, et al. Development and Reporting of Prediction Models: Guidance for Authors From Editors of Respiratory, Sleep, and Critical Care Journals. *Crit Care Med*. May 2020;48(5):623-633. doi:10.1097/CCM.0000000000004246

17. Adhikari L, Ozrazgat-Baslanti T, Ruppert M, et al. Improved predictive models for acute kidney injury with IDEA: Intraoperative Data Embedded Analytics. *PLoS One*. 2019;14(4):e0214904. doi:10.1371/journal.pone.0214904

18. Kellum JA, Lameire N, Aspelin P, et al. Kidney disease: improving global outcomes (KDIGO) acute kidney injury work group. KDIGO clinical practice guideline for acute kidney injury. *Kidney international supplements*. 2012;2(1):1-138.

19. Ozrazgat-Baslanti T, Motaei A, Islam R, et al. Development and validation of computable phenotype to identify and characterize kidney health in adult hospitalized patients. *arXiv preprint arXiv:190303149*. 2019;

20. Bihorac A, Ozrazgat-Baslanti T, Mahanna E, et al. Long-Term Outcomes for Different Forms of Stress Cardiomyopathy After Surgical Treatment for Subarachnoid Hemorrhage. *Anesth Analg*. May 2016;122(5):1594-602. doi:10.1213/ANE.0000000000001231

21. Elixhauser A, Steiner C, Harris DR, Coffey RM. Comorbidity measures for use with administrative data. *Med Care*. Jan 1998;36(1):8-27. doi:10.1097/00005650-199801000-00004

22. Wald R, Waikar SS, Liangos O, Pereira BJ, Chertow GM, Jaber BL. Acute renal failure after endovascular vs open repair of abdominal aortic aneurysm. *J Vasc Surg*. Mar 2006;43(3):460-466; discussion 466. doi:10.1016/j.jvs.2005.11.053

23. Charlson ME, Pompei P, Ales KL, MacKenzie CR. A new method of classifying prognostic comorbidity in longitudinal studies: development and validation. *J Chronic Dis*. 1987;40(5):373-83. doi:10.1016/0021-9681(87)90171-8

24. VHA V. National Drug File Reference Terminology (NDF-RT) Documentation. *US Department of Veterans Affairs*. 2012;

25. Ren Y, Loftus TJ, Datta S, et al. Performance of a Machine Learning Algorithm Using Electronic Health Record Data to Predict Postoperative Complications and Report on a Mobile Platform. *JAMA Netw Open*. May 2 2022;5(5):e2211973. doi:10.1001/jamanetworkopen.2022.11973

26. McMahan B, Moore E, Ramage D, Hampson S, y Arcas BA. Communication-efficient learning of deep networks from decentralized data. PMLR; 2017:1273-1282.

27. Li T, Sahu AK, Zaheer M, Sanjabi M, Talwalkar A, Smith V. Federated optimization in heterogeneous networks. *Proceedings of Machine learning and systems*. 2020;2:429-450.

28. Karimireddy SP, Kale S, Mohri M, Reddi S, Stich S, Suresh AT. Scaffold: Stochastic controlled averaging for federated learning. PMLR; 2020:5132-5143.

29. Dayan I, Roth HR, Zhong A, et al. Federated learning for predicting clinical outcomes in patients with COVID-19. *Nat Med*. Oct 2021;27(10):1735-1743. doi:10.1038/s41591-021-01506-3

30. Feki I, Ammar S, Kessentini Y, Muhammad K. Federated learning for COVID-19 screening from Chest X-ray images. *Appl Soft Comput*. Jul 2021;106:107330. doi:10.1016/j.asoc.2021.107330

31. Feng B, Shi J, Huang L, et al. Robustly federated learning model for identifying high-risk patients with postoperative gastric cancer recurrence. *Nat Commun*. Jan 25 2024;15(1):742. doi:10.1038/s41467-024-44946-4

**eFigure 1: Cohort selection and exclusion criteria**

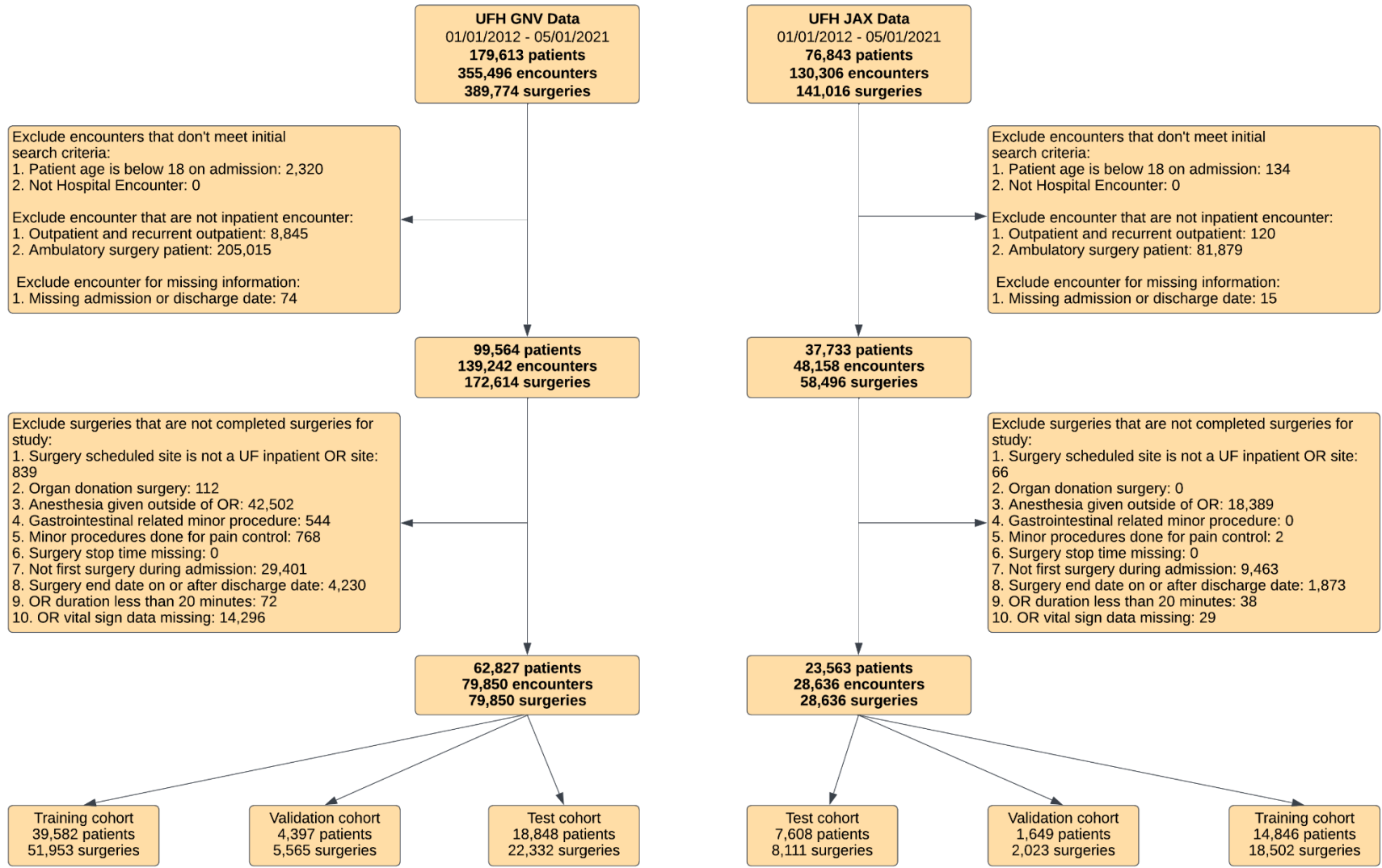

**eFigure 2: Distribution of intraoperative features used for model development in training cohort across the sites**

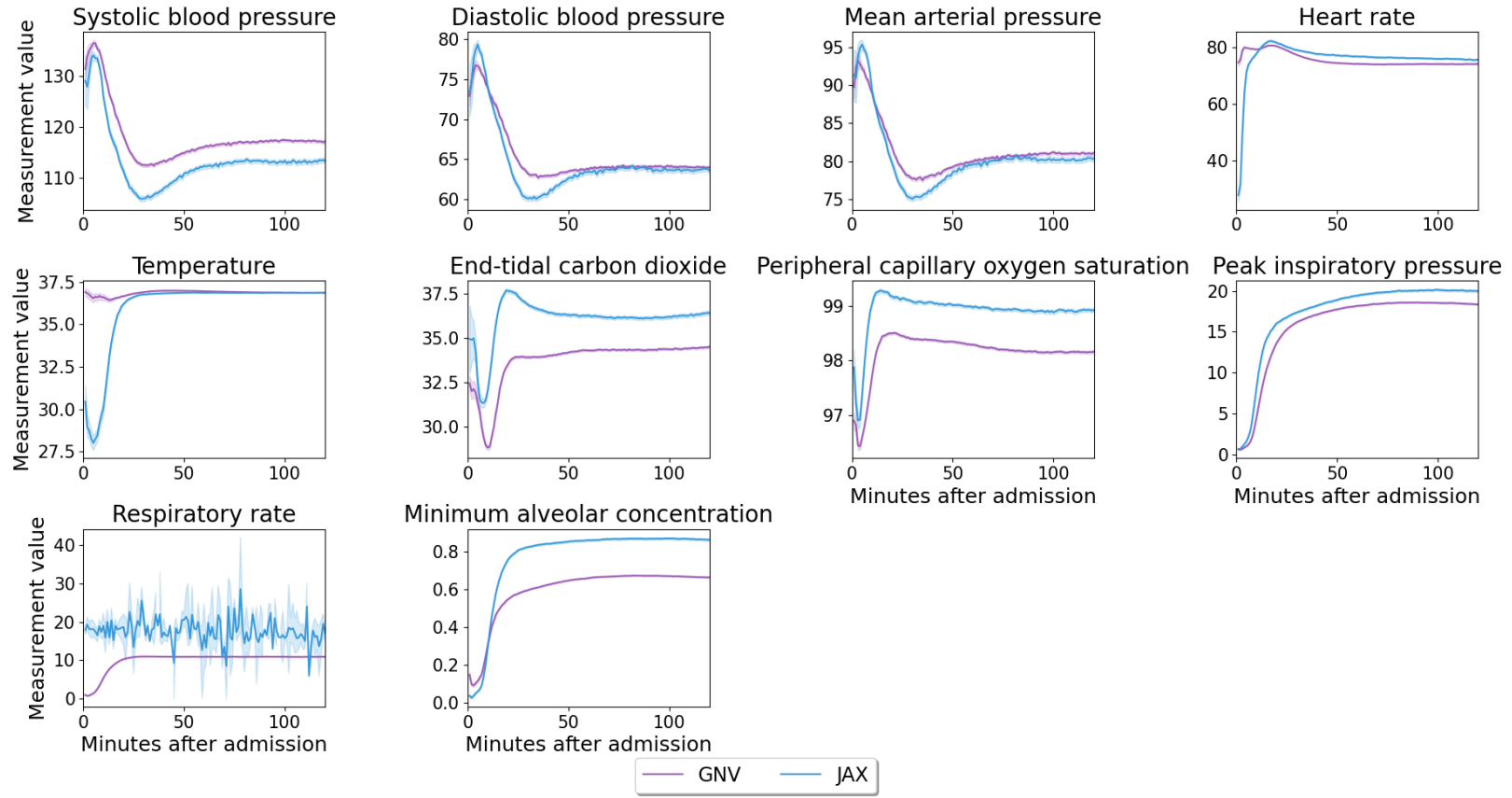

**eTable 1. Characteristics of preoperative features used for model development in training cohort across the sites**

| Features | UFH GNV | UFH JAX | P-value |
|---|---|---|---|
| Number of patients, n | 39,582 | 14,846 | |
| Number of encounters, n | 51,953 | 18,502 | |
| **Demographic information** | | | |
| Age, years, mean (SD) | 57 (17) | 52 (17) | <.001 |
| Sex, n (%) | | | |
|   Male | 26,109 (50) | 9,620 (52) | <.001 |
|   Female | 25,844 (50) | 8,882 (48) | <.001 |
| Distance of residence to hospital, km, mean (SD) | 71 (114) | 31 (119) | <.001 |
| Rural at patient residential area, n (%) | 17,980 (35) | 1,787 (10) | <.001 |
| Total population at patient residential area, mean (SD) | 20,506 (13,395) | 31,312 (16,300) | <.001 |
| Median total income at patient residential area, USD, mean (SD) | 43,254 (12,457) | 44,880 (15,656) | <.001 |
| Prevalence of residents living below poverty at patient residential area, %, mean (SD) | 20 (10) | 21 (11) | 0.28 |
| Prevalence of African American residents living below poverty at patient residential area, %, mean (SD) | 0.15 (0.15) | 0.4 (0.3) | <.001 |
| Prevalence of Hispanic residents living below poverty at patient residential area, %, mean (SD) | 0.08 (0.07) | 0.1 (0.0) | <.001 |
| County, n (%) | | | |
|   1st Rank | 12,170 (23) | 12,890 (70) | |
|   2nd Rank | 5,547 (11) | 1,593 (9) | |
|   3rd Rank | 2,321 (4) | 1,179 (6) | |
| Zip code, n (%) | | | |
|   1st Rank | 1,925 (4) | 1,698 (9) | |
|   2nd Rank | 1,139 (2) | 1,567 (8) | |
|   3rd Rank | 1,137 (2) | 1,197 (6) | |
| Marital Status, n (%) | | | |
|   Married | 25,092 (48) | 6,222 (34) | <.001 |
|   Single | 15,412 (30) | 6,675 (36) | <.001 |
|   Divorced | 8,215 (16) | 4,968 (27) | <.001 |
|   Missing | 3,234 (6) | 637 (3) | <.001 |
| Native Language Spoken, n (%) | | | |
|   English | 51,025 (98) | 17,962 (97) | <.001 |
|   Non-English | 928 (2) | 540 (3) | <.001 |
| Insurance paying the bills, n (%) | | | |
|   Medicare | 23,483 (45) | 5,602 (30) | <.001 |

| Features | UFH GNV | UFH JAX | P-value |
|---|---|---|---|
| Private | 15,840 (30) | 5,154 (28) | <.001 |
| Medicaid | 8,646 (17) | 7,493 (41) | <.001 |
| Uninsured | 3,984 (8) | 253 (1) | <.001 |
| Race, n (%)[a] | | | |
| African American | 7,166 (14) | 6,435 (35) | <.001 |
| White | 40,874 (79) | 10,904 (59) | <.001 |
| Other[b] | 3,197 (6) | 1,100 (6) | 0.32 |
| Missing | 716 (1) | 63 (0.3) | <.001 |
| Ethnicity, n (%)[a] | | | |
| Non-Hispanic | 48,946 (96) | 17,623 (96) | <.001 |
| Hispanic | 2,195 (4) | 784 (4) | 0.96 |
| Smoking Status, n (%) | | | |
| Never | 21,607 (42) | 7,130 (39) | <.001 |
| Former | 17,758 (34) | 5,102 (30) | <.001 |
| Current | 9,511 (18) | 5,549 (28) | <.001 |
| Missing | 3,077 (6) | 721 (4) | <.001 |
| Body Mass Index, median (IQR) | 28 (24, 33) | 28 (24, 33) | <.001 |
| **Surgical information, n (%)** | | | |
| Time from Admission to Surgery, days, median (IQR) | 3 (2, 21) | 4 (2, 30) | <.001 |
| Emergency admission | 20,589 (40) | 10214 (55) | <.001 |
| The admission happened at night | 24,613 (47) | 9104 (49) | <.001 |
| Admission type | | | |
| Surgery | 36,789 (85) | 15,089 (82) | <.001 |
| Medicine | 6,307 (15) | 3,413 (18) | <.001 |
| Transferred from another hospital | 8,077 (16) | 1,085 (6) | <.001 |
| Anesthesia Type | | | |
| General | 46,547 (90) | 18,112 (98) | <.001 |
| Local/regional | 5,406 (10) | 390 (2) | <.001 |
| Scheduled room is trauma room | 375 (1) | 0 (0) | <.001 |
| Scheduled post operation location is ICU | 9,999 (19) | 295 (2) | <.001 |

| Features | UFH GNV | UFH JAX | P-value |
|---|---|---|---|
| Scheduled surgery room | | | |
|   1st Rank | 3,433 (7) | 1,582 (9) | |
|   2nd Rank | 2,726 (5) | 1,462 (8) | |
|   3rd Rank | 2,716 (5) | 1,458 (8) | |
| Attending Surgeon | | | |
|   1st Rank | 8,857 (17) | 1,040 (6) | |
|   2nd Rank | 1,421(3) | 981 (5) | |
|   3rd Rank | 1,143 (2) | 736 (4) | |
| Current Procedural Terminology code of the primary procedure | | | |
|   1st Rank | 1,235 (3) | 610 (5) | |
|   2nd Rank | 881 (2) | 270 (2) | |
|   3rd Rank | 767 (2) | 246 (2) | |
| Admission Day | | | |
|   Monday | 10,076 (19) | 3,531 (19) | 0.36 |
|   Tuesday | 9,944 (19) | 2,932 (16) | <.001 |
|   Wednesday | 8,722 (17) | 3,002 (16) | 0.08 |
|   Sunday | 8,575 (17) | 3,359 (18) | <.001 |
|   Thursday | 8,656 (17) | 2,913 (16) | .003 |
|   Friday | 3,210 (6) | 1,372 (7) | <.001 |
|   Saturday | 2,770 (5) | 1,393 (8) | <.001 |
| Admission Month | | | |
|   Aug | 4,562 (9) | 1,626 (9) | 0.99 |
|   Jul | 4,438 (9) | 1,591 (9) | 0.82 |
|   Jun | 4,452 (9) | 1,582 (8) | 0.95 |
|   Oct | 4,643 (9) | 1,535 (8) | 0.008 |
|   Sep | 4,249 (8) | 1,466 (8) | 0.28 |
|   Mar | 4,116 (8) | 1,663 (9) | <.001 |
|   Jan | 4,849 (9) | 1,492 (8) | <.001 |
|   May | 3,852 (7) | 1,567 (8) | <.001 |

30| Features | UFH GNV | UFH JAX | P-value |
|---|---|---|---|
| Feb | 4,353 (8) | 1,456 (8) | 0.03 |
| Dec | 4,373 (8) | 1,490 (8) | 0.13 |
| Nov | 4,153 (8) | 1,417 (8) | 0.15 |
| Apr | 3,913 (8) | 1,617 (9) | <.001 |
| Surgery Type | | | |
| Orthopedic surgery | 10,967 (21) | 4,009 (22) | 0.11 |
| Neurosurgery | 5,947 (11) | 1,687 (9) | <.001 |
| Vascular surgery | 3,485 (7) | 763 (4) | <.001 |
| Urology | 4,055 (8) | 1,012 (6) | <.001 |
| Ear Nose Throat | 3,092 (6) | 2,374 (13) | <.001 |
| Cardio Thoracic Surgery | 2,885 (6) | 1,189 (6) | <.001 |
| Gastrointestinal Surgery | 2,726 (5) | 0 (0) | <.001 |
| OB Gynecology | 1,854 (4) | 1,484 (8) | <.001 |
| Surgical Oncology | 1,687 (3) | 0 (0) | <.001 |
| Other[c] | 15,255 (29) | 5,984 (32) | <.001 |
| **Medication within one year before admission date, n (%)** | | | |
| Number of kinds of nephrotoxic drugs received, median (IQR) | 0 (0,1) | 0 (0,1) | <.001 |
| ACE Inhibitors | 4,996 (10) | 2,173 (12) | <.001 |
| Aminoglycosides | 1,825 (4) | 1,336 (7) | <.001 |
| Antiemetic | 14,754 (28) | 5,178 (28) | 0.29 |
| Aspirin | 6,589 (13) | 2,314 (12) | 0.54 |
| Betablockers | 8,342 (16) | 2,610 (14) | <.001 |
| Bicarbonate | 1,742 (3) | 308 (2) | <.001 |
| Diuretics | 5,542 (11) | 1,930 (10) | 0.38 |
| Vasopressors or inotropes | 9,482 (18) | 3,644 (20) | <.001 |
| Statin | 4,482 (9) | 1,505 (8) | 0.04 |
| Vancomycin | 5,676 (11) | 1,789 (10) | <.001 |
| Nonsteroidal anti-inflammatory drugs | 7,195 (14) | 3,279 (18) | <.001 |

| Features | UFH GNV | UFH JAX | P-value |
|---|---|---|---|
| **Comorbidities within one year before admission date, n (%)** | | | |
| Charlson comorbidity index, median (IQR) | 2 (0, 4) | 2 (0, 4) | 0.12 |
| Alcohol or drug abuse | 6,956 (13) | 4,816 (26) | <.001 |
| Myocardial Infarction | 2,970 (6) | 1,331 (7) | <.001 |
| Congestive Heart Failure | 6,739 (13) | 2,282 (12) | 0.03 |
| Peripheral Vascular Disease | 10,006 (19) | 4,146 (22) | <.001 |
| Cerebrovascular Disease | 6,812 (13) | 2,370 (13) | 0.30 |
| Chronic Pulmonary Disease | 13,931 (27) | 5,281 (28) | <.001 |
| Cancer | 11,937 (23) | 3,672 (20) | <.001 |
| Metastatic Carcinoma | 4,190 (8) | 1,301 (7) | <.001 |
| Liver Disease | 6,093 (12) | 2,600 (14) | <.001 |
| Diabetes | 11,335 (22) | 4,685 (25) | <.001 |
| Hypertension | 7,190 (14) | 3,102 (17) | <.001 |
| Obesity | 14,061 (27) | 6,291 (34) | <.001 |
| Fluid and electrolyte disorders | 11,557 (22) | 5,523 (30) | <.001 |
| Valvular Disease | 5,460 (11) | 3,465 (19) | <.001 |
| Coagulopathy | 3,946 (8) | 1,407 (8) | 0.98 |
| Weight Loss | 6,854 (13) | 3,382 (18) | <.001 |
| Depression | 12,778 (25) | 4,535 (24) | 0.83 |
| Chronic anemia | 4,255 (8) | 2,459 (13) | <.001 |
| Chronic Kidney Disease | 8,444 (16) | 3,332 (18) | <.001 |
| **Laboratory results information** | | | |
| Have pH test with 7 days prior to surgery, n (%) | 2,493 (5) | 1,187 (6) | <.001 |
| Have pH test with 8-365 days prior to surgery, n (%) | 4,246 (8) | 1,381 (7) | 0.002 |
| Automated urinalysis, urine glucose within 7 days prior to surgery, mg/dL, n (%) | | | |
|   Missing | 39,184 (75) | 14,021 (76) | 0.33 |
|   Negative | 11,096 (21) | 3,735 (20) | <.001 |
|   Small | 931 (2) | 374 (2) | 0.05 |
|   Moderate | 382 (1) | 293 (2) | <.001 |

| Features | UFH GNV | UFH JAX | P-value |
|---|---|---|---|
|   Large | 360 (1) | 79 (0.4) | <.001 |
| Automated urinalysis, urine glucose within 8-365 days prior to surgery, mg/dL, n (%) | | | |
|   Missing | 36,131 (70) | 12,098 (65) | <.001 |
|   Negative | 13,031 (25) | 5,197 (28) | <.001 |
|   Small | 1,467 (3) | 535 (3) | 0.65 |
|   Moderate | 648 (1) | 409 (2) | <.001 |
|   Large | 676 (1) | 263 (1) | 0.23 |
| Automated urinalysis, urine protein presence within 365 days prior to surgery, mg/dL, n (%) | | | |
|   Missing | 27,278 (53) | 8,970 (48) | <.001 |
|   Negative | 10,449 (20) | 3,860 (21) | 0.03 |
|   Moderate | 9,271 (18) | 3,866 (21) | <.001 |
|   Small | 3,209 (6) | 1,049 (6) | 0.01 |
|   Large | 1,746 (3) | 757 (4) | <.001 |
| Automated urinalysis, urine red blood cell within 365 days prior to surgery, /hpf, n (%) | | | |
|   Missing | 36,435 (70) | 11,641 (63) | <.001 |
|   Negative | 9,224 (18) | 4,259 (23) | <.001 |
|   Small | 3,332 (6) | 1,439 (8) | <.001 |
|   Large | 2,453 (5) | 969 (5) | 0.005 |
|   Moderate | 509 (1) | 194 (1) | 0.44 |
| Automated urinalysis, urine hemoglobin within 7 days prior to surgery, mg/dL, n (%) | | | |
|   Missing | 39,891 (77) | 14,874 (80) | <.001 |
|   Negative | 7,143 (14) | 2,077 (11) | <.001 |
|   Small | 2,520 (5) | 699 (4) | <.001 |
|   Moderate | 1,342 (3) | 463 (3) | 0.57 |
|   Large | 1,057 (2) | 389 (2) | 0.60 |
| Automated urinalysis, urine hemoglobin within 8-365 days prior to surgery, mg/dL, n (%) | | | |
|   Missing | 38,935 (75) | 13,161 (71) | <.001 |

| Features | UFH GNV | UFH JAX | P-value |
|---|---|---|---|
| Negative | 6,690 (13) | 2,934 (16) | <.001 |
| Small | 2,839 (5) | 1,017 (5) | 0.88 |
| Large | 1,807 (3) | 697 (4) | 0.07 |
| Moderate | 1,682 (3) | 693 (4) | 0.001 |
| Hemoglobin within 7 days prior to surgery, g/dl, median (IQR) | | | |
| Minimum | 12.60 (10.8, 13.9) | 12.0 (10.3, 13.5) | <.001 |
| Maximum | 13.2 (11.7, 14.4) | 12.8 (11.3, 14.0) | <.001 |
| Average | 12.9 (11.2, 14.1) | 12.4 (10.8, 13.7) | <.001 |
| Variance | 0 (0, 0.3) | 0 (0, 0.5) | <.001 |
| Count | 1 (0, 2) | 1 (0, 2) | <.001 |
| Hemoglobin within 8-365 days prior to surgery, g/dl, median (IQR) | | | |
| Minimum | 12.0 (9.7, 13.6) | 11.6 (9.5, 13.2) | <.001 |
| Maximum | 13.7 (12.5, 14.8) | 13.3 (12.1, 14.4) | <.001 |
| Average | 12.7 (11.0, 13.9) | 12.3 (10.8, 13.6) | <.001 |
| Variance | 0.4 (0, 1.3) | 0.4 (0, 1.3) | <.001 |
| Count | 1 (0, 4) | 1 (0, 4) | <.001 |
| Glucose in blood within 7 days prior to surgery, mg/dL, median (IQR) | | | |
| Minimum | 100 (88, 119) | 99 (87, 119) | 0.04 |
| Maximum | 118 (98, 155) | 122 (100, 162) | <.001 |
| Average | 110 (95, 135) | 111 (96, 137) | <.001 |
| Variance | 0 (0, 283) | 22 (0, 374) | <.001 |
| Count | 1 (0, 2) | 1 (0, 3) | <.001 |
| Glucose in blood within 8-365 days prior to surgery, mg/dL, median (IQR) | | | |
| Minimum | 89 (79, 101) | 88 (77, 99) | <.001 |
| Maximum | 128 (101, 180) | 127 (101, 185) | 0.64 |
| Average | 108 (95, 128) | 106 (94, 127) | <.001 |
| Variance | 168 (0, 669) | 199 (5, 742) | <.001 |

| Features | UFH GNV | UFH JAX | P-value |
|---|---|---|---|
| Count | 1 (0, 5) | 1 (0, 6) | <.001 |
| Urea nitrogen in blood within 7 days prior to surgery, mg/dL, median (IQR) | | | |
|   Minimum | 14.0 (10.0, 19.0) | 12.0 (9.0, 17.0) | <.001 |
|   Maximum | 16.0 (12.0, 21.0) | 14.0 (11.0, 19.0) | <.001 |
|   Average | 15.0 (11.0, 20.0) | 13.0 (10.0, 18.0) | <.001 |
|   Variance | 0.0 (0.0, 2.9) | 0.0 (0.0, 4.3) | <.001 |
|   Count | 1 (0, 2) | 1 (0, 2) | <.001 |
| Urea nitrogen in blood within 8-365 days prior to surgery, mg/dL, median (IQR) | | | |
|   Minimum | 12.0 (8.0, 16.0) | 11.0 (8.0, 15.0) | <.001 |
|   Maximum | 18.0 (13.0, 24.0) | 16.0 (12.0, 22.0) | <.001 |
|   Average | 14.7 (11.1, 19.0) | 13.5 (10.5, 17.8) | <.001 |
|   Variance | 4.3 (0.0, 15.5) | 4.5 (0.0, 15.9) | <.001 |
|   Count | 1 (0, 4) | 1 (0, 4) | <.001 |
| Serum creatinine within 7 days prior to surgery, mg/dL, median (IQR) | | | |
|   Minimum | 0.8 (0.7, 1.0) | 0.8 (0.7, 1.0) | <.001 |
|   Maximum | 0.9 (0.7, 1.1) | 0.9 (0.7, 1.1) | 0.03 |
|   Average | 0.9 (0.7, 1.1) | 0.9 (0.7, 1.1) | 0.009 |
|   Variance | 0.0 (0.0, 0.0) | 0.0 (0.0, 0.0) | <.001 |
|   Count | 1 (0, 2) | 1 (0, 2) | <.001 |
| Serum creatinine within 8-365 days prior to surgery, mg/dL, median (IQR) | | | |
|   Minimum | 0.8 (0.6, 1.0) | 0.8 (0.6, 0.9) | <.001 |
|   Maximum | 1.0 (0.8, 1.2) | 0.9 (0.8, 1.2) | <.001 |
|   Average | 0.9 (0.7, 1.1) | 0.8 (0.7, 1.0) | <.001 |
|   Variance | 0.0 (0.0, 0.0) | 0.0 (0.0, 0.0) | <.001 |
|   Count | 1 (0, 4) | 1 (0, 4) | <.001 |
| Serum Calcium within 7 days prior to surgery, mmol/L, median (IQR) | | | |
|   Minimum | 9.1 (8.6, 9.6) | 8.9 (8.4, 9.4) | <.001 |

35| Features | UFH GNV | UFH JAX | P-value |
|---|---|---|---|
| Maximum | 9.4 (9.0, 9.7) | 9.2 (8.9, 9.6) | <.001 |
| Average | 9.2 (8.8, 9.6) | 9.1 (8.7, 9.4) | <.001 |
| Variance | 0.0 (0.0, 0.1) | 0.0 (0.0, 0.1) | <.001 |
| Count | 1 (0, 2) | 1 (0, 2) | <.001 |
| Serum Calcium within 8-365 days prior to surgery, mmol/L, median (IQR) | | | |
| Minimum | 8.9 (8.2, 9.4) | 8.9 (8.3, 9.4) | 0.85 |
| Maximum | 9.6 (9.3, 9.9) | 9.6 (9.3, 9.9) | <.001 |
| Average | 9.2 (8.8, 9.6) | 9.2 (8.9, 9.5) | 0.05 |
| Variance | 0.1 (0.0, 0.2) | 0.1 (0.0, 0.2) | 0.17 |
| Count | 1 (0, 4) | 1 (0, 4) | <.001 |
| Serum Sodium within 7 days prior to surgery, mmol/L, median (IQR) | | | |
| Minimum | 138 (136, 140) | 138 (136, 140) | 0.51 |
| Maximum | 140 (138, 142) | 140 (138, 142) | <.001 |
| Average | 139 (137, 141) | 139 (137, 141) | 0.003 |
| Variance | 0 (0, 2) | 0 (0, 3) | <.001 |
| Count | 1 (0, 2) | 1 (0, 2) | <.001 |
| Serum Sodium within 8-365 days prior to surgery, mmol/L, median (IQR) | | | |
| Minimum | 137 (134, 140) | 137 (134, 140) | 0.02 |
| Maximum | 141 (139, 143) | 141 (140, 143) | <.001 |
| Average | 139 (137, 140) | 139 (137, 141) | <.001 |
| Variance | 2 (0, 6) | 3 (0, 8) | <.001 |
| Count | 1 (0, 4) | 1 (0, 4) | <.001 |
| Urea nitrogen-Creatinine ratio within 7 days prior to surgery, median (IQR) | | | |
| Minimum | 16.1 (12.4, 20.8) | 14.1 (10.6, 18.4) | <.001 |
| Maximum | 18.0 (14.1, 23.3) | 16.2 (12.5, 21.2) | <.001 |
| Average | 17.1 (13.4, 22.0) | 15.2 (11.6, 19.7) | <.001 |
| Variance | 0.0 (0.0, 2.9) | 0.0 (0.0, 3.8) | <.001 |
| Count | 0 (0, 0) | 1 (0, 2) | <.001 |

| Features | UFH GNV | UFH JAX | P-value |
|---|---|---|---|
| Urea nitrogen-Creatinine ratio within 8-365 days prior to surgery, median (IQR) | | | |
|   Minimum | 14.2 (10.7, 18.8) | 12.5 (9.2, 16.3) | <.001 |
|   Maximum | 20.0 (15.4, 26.1) | 18.9 (14.6, 24.6) | <.001 |
|   Average | 17.2 (13.7, 21.7) | 15.7 (12.5, 19.8) | <.001 |
|   Variance | 2.9 (0.0, 14.7) | 5.1 (0.0, 17.0) | <.001 |
|   Count | 0 (0, 0) | 1 (0, 4) | <.001 |
| Potassium in serum within 7 days prior to surgery, mmol/L, median (IQR) | | | |
|   Minimum | 3.9 (3.6, 4.2) | 3.8 (3.5, 4.1) | <.001 |
|   Maximum | 4.2 (3.9, 4.5) | 4.1 (3.8, 4.4) | <.001 |
|   Average | 4.0 (3.8, 4.3) | 4.0 (3.7, 4.2) | <.001 |
|   Variance | 0.0 (0.0, 0.1) | 0.0 (0.0, 0.1) | 0.03 |
|   Count | 1 (0, 2) | 1 (0, 2) | <.001 |
| Potassium in serum within 8-365 days prior to surgery, mmol/L, median (IQR) | | | |
|   Minimum | 3.8 (3.4, 4.1) | 3.8 (3.4, 4.1) | 0.89 |
|   Maximum | 4.4 (4.1, 4.8) | 4.4 (4.1, 4.8) | <.001 |
|   Average | 4.1 (3.9, 4.3) | 4.1 (3.8, 4.3) | 0.06 |
|   Variance | 0.1 (0.0, 0.2) | 0.1 (0.0, 0.2) | <.001 |
|   Count | 1 (0, 4) | 1 (0, 4) | <.001 |
| Chloride in Serum within 7 days prior to surgery, mmol/L, mmol/L, median (IQR) | | | |
|   Minimum | 101 (98, 103) | 100 (97, 102) | <.001 |
|   Maximum | 102 (100, 105) | 102 (99, 104) | <.001 |
|   Average | 102 (99, 104) | 101 (98, 103) | <.001 |
|   Variance | 0 (0, 2) | 0 (0, 4) | <.001 |
|   Count | 1 (0, 2) | 1 (0, 2) | <.001 |
| Chloride in Serum within 8-365 days prior to surgery, mmol/L, mmol/L, median (IQR) | | | |
|   Minimum | 100 (97, 102) | 99 (96, 102) | <.001 |
|   Maximum | 104 (101, 107) | 104 (101, 106) | <.001 |

| Features | UFH GNV | UFH JAX | P-value |
|---|---|---|---|
| Average | 102 (100, 104) | 101 (99, 103) | <.001 |
| Variance | 2 (0, 8) | 4 (0, 10) | <.001 |
| Count | 1 (0, 4) | 1 (0, 4) | <.001 |
| Serum CO2 within 7 days prior to surgery, mmol/L, median (IQR) | | | |
| Minimum | 24 (22, 26) | 23 (21, 25) | <.001 |
| Maximum | 26 (24, 28) | 25 (23, 27) | <.001 |
| Average | 25 (23, 27) | 24 (22, 26) | <.001 |
| Variance | 0 (0, 2) | 0 (0, 2) | 0.41 |
| Count | 1 (0, 2) | 1 (0, 2) | <.001 |
| Serum CO2 within 8-365 days prior to surgery, mmol/L, median (IQR) | | | |
| Minimum | 23 (21, 26) | 23 (20, 25) | <.001 |
| Maximum | 27 (25, 29) | 26 (25, 28) | <.001 |
| Average | 25 (23, 27) | 24 (23, 26) | <.001 |
| Variance | 2 (0, 6) | 2 (0, 6) | <.001 |
| Count | 1 (0, 4) | 1 (0, 4) | <.001 |
| Anion gap in blood within 7 days prior to surgery, mmol/L, median (IQR) | | | |
| Minimum | 12 (9, 14) | 13 (11, 15) | <.001 |
| Maximum | 13 (11, 16) | 15 (13, 17) | <.001 |
| Average | 12 (10, 15) | 14 (12, 15) | <.001 |
| Variance | 0 (0, 2) | 0 (0, 2) | 0.40 |
| Count | 1 (0, 1) | 1 (0, 2) | <.001 |
| Anion gap in blood within 8-365 days prior to surgery, mmol/L, median (IQR) | | | |
| Minimum | 10 (7, 12) | 12 (10, 14) | <.001 |
| Maximum | 14 (12, 17) | 15 (14, 18) | <.001 |
| Average | 12 (10, 14) | 13 (12, 15) | <.001 |
| Variance | 2 (0, 6) | 2 (0, 5) | 0.10 |
| Count | 0 (0, 2) | 1 (0, 3) | <.001 |

| Features | UFH GNV | UFH JAX | P-value |
|---|---|---|---|
| Band form neutrophils/100 leukocytes in blood within 7 days prior to surgery, %, median (IQR) | | | |
|   Minimum | 5 (2, 12) | 4 (1, 9) | <.001 |
|   Maximum | 8 (3, 19) | 6 (2, 15) | <.001 |
|   Average | 7 (3, 16) | 5 (2, 12) | <.001 |
|   Variance | 0 (0, 1) | 0 (0, 2) | 0.40 |
|   Count | 0 (0, 0) | 0 (0, 0) | <.001 |
| Band form neutrophils/100 leukocytes in blood within 8-365 days prior to surgery, %, median (IQR) | | | |
|   Minimum | 3 (1, 7) | 2 (1, 4) | <.001 |
|   Maximum | 8 (3, 19) | 7 (2, 18) | <.001 |
|   Average | 5 (2, 11) | 4 (2, 10) | <.001 |
|   Variance | 0 (0, 25) | 0 (0, 29) | 0.03 |
|   Count | 0 (0, 0) | 0 (0, 0) | <.001 |
| White Blood Cell in blood within 7 days prior to surgery, thou/uL, median (IQR) | | | |
|   Minimum | 7.6 (6.0, 10.0) | 8.1 (6.2, 11.0) | <.001 |
|   Maximum | 8.7 (6.6, 12.0) | 9.6 (7.1, 13.6) | <.001 |
|   Average | 8.2 (6.4, 10.9) | 8.9 (6.7, 12.2) | <.001 |
|   Variance | 0.0 (0.0, 1.0) | 0.0 (0.0, 1.6) | <.001 |
|   Count | 1 (0, 2) | 1 (0, 2) | <.001 |
| White Blood Cell in blood within 8-365 days prior to surgery, thou/uL, median (IQR) | | | |
|   Minimum | 6.3 (4.9, 7.9) | 6.3 (4.9, 8.0) | 0.93 |
|   Maximum | 9.2 (7.0, 12.9) | 9.2 (6.8, 13.0) | 0.19 |
|   Average | 7.7 (6.2, 9.7) | 7.8 (6.1, 9.8) | 0.77 |
|   Variance | 0.9 (0.0, 4.7) | 1.0 (0.0, 5.1) | <.001 |
|   Count | 1 (0, 4) | 1 (0, 4) | <.001 |
| Serum Red Blood Cell within 7 days prior to surgery, Million/uL, median (IQR) | | | |
|   Minimum | 4.2 (3.7, 4.7) | 4.1 (3.6, 4.6) | <.001 |
|   Maximum | 4.4 (4.0, 4.8) | 4.4 (3.9, 4.8) | <.001 |

| Features | UFH GNV | UFH JAX | P-value |
|---|---|---|---|
| Average | 4.3 (3.8, 4.7) | 4.2 (3.8, 4.7) | <.001 |
| Variance | 0.0 (0.0, 0.03) | 0.0 (0.0, 0.05) | <.001 |
| Count | 1 (0, 2) | 1 (0, 2) | <.001 |
| Serum Red Blood Cell within 8-365 days prior to surgery, Million/uL, median (IQR) | | | |
| Minimum | 4.1 (3.4, 4.6) | 4.0 (3.4, 4.5) | 0.003 |
| Maximum | 4.6 (4.2, 4.9) | 4.6 (4.2, 4.9) | <.001 |
| Average | 4.3 (3.8, 4.7) | 4.2 (3.8, 4.6) | 0.006 |
| Variance | 0.0 (0.0, 0.1) | 0.0 (0.0, 0.2) | <.001 |
| Count | 1 (0, 4) | 1 (0, 4) | <.001 |
| Hematocrit in blood within 7 days prior to surgery, %, median (IQR) | | | |
| Minimum | 38.1 (33.0, 41.9) | 36.7 (31.9, 40.5) | <.001 |
| Maximum | 39.8 (35.8, 43.3) | 38.5 (34.6, 41.9) | <.001 |
| Average | 38.9 (34.4, 42.4) | 37.5 (33.3, 41.0) | <.001 |
| Variance | 0.0 (0.0, 3.0) | 0.0 (0.0, 3.6) | <.001 |
| Count | 1 (0, 2) | 1 (0, 2) | <.001 |
| Hematocrit in blood within 8-365 days prior to surgery, %, median (IQR) | | | |
| Minimum | 36.4 (30.0, 41.0) | 35.6 (29.6, 39.9) | <.001 |
| Maximum | 41.4 (38.2, 44.5) | 40.3 (37.1, 43.4) | <.001 |
| Average | 38.4 (34.0, 42.0) | 37.5 (33.2, 41.0) | <.001 |
| Variance | 3.4 (0.0, 12.2) | 3.4 (0.0, 11.7) | 0.03 |
| Count | 1 (0, 4) | 1 (0, 4) | <.001 |
| The amount of hemoglobin relative to the size of the cell in blood within 7 days prior to surgery, g/dL, median (IQR) | | | |
| Minimum | 32.7 (31.7, 33.5) | 32.8 (31.8, 33.7) | <.001 |
| Maximum | 33.5 (32.6, 34.2) | 33.2 (32.3, 34.1) | <.001 |
| Average | 33.1 (32.2, 33.8) | 33.0 (32.1, 33.9) | 0.17 |
| Variance | 0.1 (0.0, 0.4) | 0.0 (0.0, 0.2) | <.001 |
| Count | 2 (0, 2) | 1 (0, 2) | <.001 |

| Features | UFH GNV | UFH JAX | P-value |
|---|---|---|---|
| The amount of hemoglobin relative to the size of the cell in blood within 8-365 days prior to surgery, g/dL, median (IQR) | | | |
|   Minimum | 32.2 (31.1, 33.2) | 32.2 (31.1, 33.1) | 0.20 |
|   Maximum | 33.7 (32.9, 34.5) | 33.4 (32.5, 34.3) | <.001 |
|   Average | 33.0 (32.1, 33.7) | 32.8 (31.9, 33.6) | <.001 |
|   Variance | 0.3 (0.1, 0.6) | 0.2 (0.0, 0.6) | <.001 |
|   Count | 1 (0, 6) | 1 (0, 4) | <.001 |
| Red cell distribution width in Blood within 7 days prior to surgery, %, median (IQR) | | | |
|   Minimum | 14.1 (13.3, 15.1) | 13.5 (12.8, 14.7) | <.001 |
|   Maximum | 14.4 (13.5, 15.5) | 13.7 (12.9, 15.0) | <.001 |
|   Average | 14.2 (13.4, 15.3) | 13.6 (12.8, 14.8) | <.001 |
|   Variance | 0.0 (0.0, 0.0) | 0.0 (0.0, 0.0) | <.001 |
|   Count | 1 (0, 2) | 1 (0, 2) | <.001 |
| Red cell distribution width in Blood within 8-365 days prior to surgery, %, median (IQR) | | | |
|   Minimum | 13.8 (13.1, 14.7) | 13.5 (12.8, 14.6) | <.001 |
|   Maximum | 14.8 (13.8, 16.3) | 14.5 (13.4, 16.1) | <.001 |
|   Average | 14.3 (13.5, 15.4) | 14.0 (13.2, 15.3) | <.001 |
|   Variance | 0.1 (0.0, 0.6) | 0.1 (0.0, 0.5) | 0.04 |
|   Count | 1 (0, 3) | 1 (0, 4) | <.001 |
| Platelet in blood within 7 days prior to surgery, thou/uL, median (IQR) | | | |
|   Minimum | 225 (177, 281) | 237 (186, 294) | <.001 |
|   Maximum | 242 (194, 303) | 258 (207, 320) | <.001 |
|   Average | 233 (186, 291) | 247 (197, 305) | <.001 |
|   Variance | 0 (0, 242) | 0 (0, 427) | <.001 |
|   Count | 1 (0, 2) | 1 (0, 2) | <.001 |
| Platelet in blood within 8-365 days prior to surgery, thou/uL, median (IQR) | | | |
|   Minimum | 203 (155, 255) | 217 (166, 270) | <.001 |

| Features | UFH GNV | UFH JAX | P-value |
|---|---|---|---|
| Maximum | 266 (210, 345) | 283 (227, 360) | <.001 |
| Average | 235 (189, 290) | 251 (202, 304) | <.001 |
| Variance | 333 (0, 1850) | 420 (0, 2025) | <.001 |
| Count | 1 (0, 4) | 1 (0, 4) | <.001 |
| Mean platelet volume within 7 days prior to surgery, fL, median (IQR) | | | |
| Minimum | 7.9 (7.3, 8.7) | 10.2 (9.6, 10.9) | <.001 |
| Maximum | 8.3 (7.6, 9.2) | 10.4 (9.8, 11.1) | <.001 |
| Average | 8.1 (7.5, 8.9) | 10.3 (9.7, 11.0) | <.001 |
| Variance | 0.0 (0.0, 0.1) | 0.0 (0.0, 0.0) | <.001 |
| Count | 1 (0, 2) | 1 (0, 2) | <.001 |
| Mean platelet volume within 8-365 days prior to surgery, fL, median (IQR) | | | |
| Minimum | 7.7 (7.1, 8.5) | 9.9 (9.3, 10.7) | <.001 |
| Maximum | 8.9 (8.1, 9.8) | 10.8 (10.1, 11.5) | <.001 |
| Average | 8.2 (7.6, 9.0) | 10.3 (9.7, 11.0) | <.001 |
| Variance | 0.1 (0.0, 0.5) | 0.1 (0.0, 0.2) | <.001 |
| Count | 1 (0, 3) | 1 (0, 4) | <.001 |
| Mean Corpuscular Volume in blood within 7 days prior to surgery, fL, median (IQR) | | | |
| Minimum | 90.1 (86.1, 93.9) | 88.6 (84.5, 92.3) | <.001 |
| Maximum | 90.8 (86.9, 94.7) | 89.5 (85.4, 93.3) | <.001 |
| Average | 90.5 (86.5, 94.2) | 89.0 (85.0, 92.7) | <.001 |
| Variance | 0.0 (0.0, 0.4) | 0.0 (0.0, 0.7) | <.001 |
| Count | 1 (0, 2) | 1 (0, 2) | <.001 |
| Mean Corpuscular Volume in blood within 8-365 days prior to surgery, fL, median (IQR) | | | |
| Minimum | 89.0 (84.8, 92.9) | 87.7 (83.2, 91.5) | <.001 |
| Maximum | 92.0 (88.0, 96.1) | 90.8 (86.6, 94.9) | <.001 |
| Average | 90.4 (86.6, 94.2) | 89.2 (85.1, 93.0) | <.001 |
| Variance | 0.8 (0.0, 3.3) | 1.2 (0.0, 3.9) | <.001 |
| Count | 1 (0, 4) | 1 (0, 4) | <.001 |

42| Features | UFH GNV | UFH JAX | P-value |
|---|---|---|---|
| Mean Corpuscular Hemoglobin in blood within 7 days prior to surgery, fL, median (IQR) | | | |
|   Minimum | 29.8 (28.2, 31.2) | 29.4 (27.7, 30.8) | <.001 |
|   Maximum | 30.1 (28.6, 31.5) | 29.7 (28.0, 31.1) | <.001 |
|   Average | 29.9 (28.4, 31.3) | 29.5 (27.8, 31.0) | <.001 |
|   Variance | 0.0 (0.0, 0.1) | 0.0 (0.0, 0.1) | <.001 |
|   Count | 1 (0, 2) | 1 (0, 2) | <.001 |
| Mean Corpuscular Hemoglobin in blood within 8-365 days prior to surgery, fL, median (IQR) | | | |
|   Minimum | 29.3 (27.6, 30.8) | 28.9 (26.9, 30.4) | <.001 |
|   Maximum | 30.5 (28.9, 31.9) | 29.9 (28.2, 31.4) | <.001 |
|   Average | 29.9 (28.3, 31.3) | 29.4 (27.6, 30.8) | <.001 |
|   Variance | 0.1 (0.0, 0.5) | 0.1 (0.0, 0.4) | 0.39 |
|   Count | 1 (0, 4) | 1 (0, 4) | <.001 |
| Serum Lactate within 7 days prior to surgery, mmol/L, median (IQR) | | | |
|   Minimum | 1.2 (0.9, 1.9) | 1.6 (1.1, 2.3) | <.001 |
|   Maximum | 1.6 (1.1, 2.6) | 1.7 (1.2, 2.7) | <.001 |
|   Average | 1.5 (1.0, 2.2) | 1.7 (1.2, 2.5) | <.001 |
|   Variance | 0.0 (0.0, 0.1) | 0.0 (0.0, 0.0) | <.001 |
|   Count | 0 (0, 0) | 0 (0, 0) | <.001 |
| Serum Lactate within 8-365 days prior to surgery, mmol/L, median (IQR) | | | |
|   Minimum | 0.9 (0.7, 1.3) | 1.2 (0.9, 1.7) | <.001 |
|   Maximum | 1.7 (1.2, 2.7) | 1.8 (1.2, 2.8) | <.001 |
|   Average | 1.3 (1.0, 1.8) | 1.5 (1.1, 2.2) | <.001 |
|   Variance | 0.1 (0.0, 0.4) | 0.0 (0.0, 0.4) | <.001 |
|   Count | 0 (0, 0) | 0 (0, 0) | <.001 |
| Serum Alanine aminotransferase within 7 days prior to surgery, U/L, median (IQR) | | | |
|   Minimum | 17 (12, 28) | 17 (11, 29) | 0.38 |
|   Maximum | 19 (12, 30) | 19 (12, 33) | 0.51 |

| Features | UFH GNV | UFH JAX | P-value |
|---|---|---|---|
| Average | 18 (12, 29) | 18 (12, 31) | 0.92 |
| Variance | 0 (0, 0) | 0 (0, 0) | <.001 |
| Count | 0 (0, 1) | 0 (0, 1) | <.001 |
| Serum Alanine aminotransferase within 8-365 days prior to surgery, U/L, median (IQR) | | | |
| Minimum | 15 (10, 22) | 14 (10, 22) | 0.83 |
| Maximum | 22 (14, 38) | 20 (14, 33) | <.001 |
| Average | 18 (13, 28) | 17 (12, 27) | <.001 |
| Variance | 2 (0, 40) | 1 (0, 24) | <.001 |
| Count | 0 (0, 1) | 0 (0, 2) | <.001 |
| Serum Albumin within 7 days prior to surgery, g/dL, median (IQR) | | | |
| Minimum | 3.9 (3.3, 4.3) | 3.6 (3.1, 4.1) | <.001 |
| Maximum | 4.0 (3.4, 4.3) | 3.8 (3.3, 4.2) | <.001 |
| Average | 3.9 (3.3, 4.3) | 3.7 (3.2, 4.1) | <.001 |
| Variance | 0.0 (0.0, 0.0) | 0.0 (0.0, 0.0) | <.001 |
| Count | 0 (0, 1) | 0 (0, 1) | <.001 |
| Serum Albumin within 8-365 days prior to surgery, g/dL, median (IQR) | | | |
| Minimum | 3.8 (3.2, 4.2) | 3.9 (3.3, 4.2) | 0.02 |
| Maximum | 4.2 (3.9, 4.5) | 4.2 (3.9, 4.4) | <.001 |
| Average | 4.0 (3.6, 4.3) | 4.0 (3.6, 4.3) | 0.65 |
| Variance | 0.0 (0.0, 0.1) | 0.0 (0.0, 0.1) | 0.002 |
| Count | 0 (0, 1) | 0 (0, 2) | <.001 |
| Serum Aspartate aminotransferase within 7 days prior to surgery, U/L, median (IQR) | | | |
| Minimum | 21 (16, 30) | 20 (15, 30) | <.001 |
| Maximum | 22 (17, 33) | 22 (16, 35) | <.001 |
| Average | 22 (16, 32) | 21 (15, 33) | <.001 |
| Variance | 0 (0, 0) | 0 (0, 0) | <.001 |
| Count | 0 (0, 1) | 0 (0, 1) | <.001 |

| Features | UFH GNV | UFH JAX | P-value |
|---|---|---|---|
| Serum Aspartate aminotransferase within 8-365 days prior to surgery, U/L, median (IQR) | | | |
|   Minimum | 18 (14, 23) | 17 (13, 23) | <.001 |
|   Maximum | 24 (18, 38) | 22 (17, 33) | 0.01 |
|   Average | 21 (16, 29) | 20 (15, 27) | <.001 |
|   Variance | 2 (0, 41) | 2 (0, 24) | <.001 |
|   Count | 0 (0, 1) | 0 (0, 2) | <.001 |
| Serum Bilirubin direct within 7 days prior to surgery, mg/dL, median (IQR) | | | |
|   Minimum | 0.2 (0.2, 0.2) | 0.1 (0.1, 0.2) | <.001 |
|   Maximum | 0.2 (0.2, 0.2) | 0.2 (0.1, 0.2) | <.001 |
|   Average | 0.2 (0.2, 0.2) | 0.2 (0.1, 0.2) | <.001 |
|   Variance | 0.0 (0.0, 0.0) | 0.0 (0.0, 0.0) | 0.09 |
|   Count | 0 (0, 0) | 0 (0, 0) | <.001 |
| Serum Bilirubin direct within 8-365 days prior to surgery, mg/dL, median (IQR) | | | |
|   Minimum | 0.2 (0.1, 0.2) | 0.1 (0.1, 0.2) | <.001 |
|   Maximum | 0.2 (0.2, 0.2) | 0.1 (0.1, 0.2) | <.001 |
|   Average | 0.2 (0.2, 0.2) | 0.1 (0.1, 0.2) | <.001 |
|   Variance | 0.0 (0.0, 0.0) | 0.0 (0.0, 0.0) | <.001 |
|   Count | 0 (0, 0) | 0 (0, 0) | <.001 |
| Serum C reactive protein within 7 days prior to surgery, mg/L, median (IQR) | | | |
|   Minimum | 35.5 (6.4, 107.4) | 58.6 (12.0, 144.2) | <.001 |
|   Maximum | 39.4 (6.8, 114.4) | 60.6 (12.3, 147.6) | <.001 |
|   Average | 38.3 (6.7, 111.0) | 60.1 (12.2, 144.9) | <.001 |
|   Variance | 0.0 (0.0, 0.0) | 0.0 (0.0, 0.0) | 0.005 |
|   Count | 0 (0, 0) | 0 (0, 0) | <.001 |
| Serum C reactive protein within 8-365 days prior to surgery, mg/L, median (IQR) | | | |
|   Minimum | 9.2 (2.6, 44.8) | 11.7 (3.4, 55.2) | <.001 |
|   Maximum | 34.9 (5.3, 117.5) | 26.6 (5.3, 104.3) | 0.01 |

45| Features | UFH GNV | UFH JAX | P-value |
|---|---|---|---|
| Average | 23.8 (4.7, 79.1) | 19.3 (4.8, 82.0) | 0.86 |
| Variance | 0.0 (0.0, 521.7) | 0.0 (0.0, 20.6) | <.001 |
| Count | 0 (0, 0) | 0 (0, 0) | <.001 |
| Serum INR within 7 days prior to surgery, median (IQR) | | | |
| Minimum | 1.1 (1.0, 1.2) | 1.1 (1.0, 1.1) | <.001 |
| Maximum | 1.1 (1.0, 1.2) | 1.1 (1.0, 1.2) | <.001 |
| Average | 1.1 (1.0, 1.2) | 1.1 (1.0, 1.2) | <.001 |
| Variance | 0.0 (0.0, 0.0) | 0.0 (0.0, 0.0) | <.001 |
| Count | 0 (0, 1) | 1 (0, 1) | <.001 |
| Serum INR within 8-365 days prior to surgery, median (IQR) | | | |
| Minimum | 1.0 (1.0, 1.1) | 1.0 (1.0, 1.1) | <.001 |
| Maximum | 1.1 (1.0, 1.3) | 1.1 (1.0, 1.2) | <.001 |
| Average | 1.1 (1.0, 1.2) | 1.1 (1.0, 1.2) | <.001 |
| Variance | 0.0 (0.0, 0.0) | 0.0 (0.0, 0.0) | 0.09 |
| Count | 0 (0, 1) | 0 (0, 1) | 0.03 |
| Erythrocyte sedimentation rate within 7 days prior to surgery, mm/h, median (IQR) | | | |
| Minimum | 47 (22, 83) | 56 (28, 86) | <.001 |
| Maximum | 48 (23, 85) | 57 (29, 87) | <.001 |
| Average | 48 (22, 84) | 56 (29, 87) | <.001 |
| Variance | 0 (0, 0) | 0 (0, 0) | 0.06 |
| Count | 0 (0, 0) | 0 (0, 0) | <.001 |
| Erythrocyte sedimentation rate within 8-365 days prior to surgery, mm/h, median (IQR) | | | |
| Minimum | 28 (11, 55) | 31 (13, 66) | <.001 |
| Maximum | 40 (17, 78) | 42 (16, 86) | 0.12 |
| Average | 35 (16, 65) | 39 (15, 76) | .004 |
| Variance | 0 (0, 53) | 0 (0, 40) | 0.33 |
| Count | 0 (0, 0) | 0 (0, 0) | <.001 |
| Serum Troponin I within 7 days prior to surgery, ng/mL, median (IQR) | | | |

| Features | UFH GNV | UFH JAX | P-value |
|---|---|---|---|
| Minimum | 0.03 (0.03, 0.1) | 0.05 (0.05, 0.05) | <.001 |
| Maximum | 0.04 (0.03, 0.1) | 0.05 (0.05, 0.05) | <.001 |
| Average | 0.04 (0.03, 0.1) | 0.05 (0.05, 0.05) | <.001 |
| Variance | 0.0 (0.0, 0.0) | 0.0 (0.0, 0.0) | <.001 |
| Count | 0 (0, 0) | 0 (0, 0) | <.001 |
| Serum Troponin I within 8-365 days prior to surgery, ng/mL, median (IQR) | | | |
| Minimum | 0.03 (0.03, 0.1) | 0.05 (0.05, 0.05) | <.001 |
| Maximum | 0.04 (0.03, 0.1) | 0.05 (0.05, 0.05) | <.001 |
| Average | 0.04 (0.03, 0.1) | 0.05 (0.05, 0.05) | <.001 |
| Variance | 0.0 (0.0, 0.0) | 0.0 (0.0, 0.0) | <.001 |
| Count | 0 (0, 0) | 0 (0, 0) | <.001 |
| Serum Troponin T within 7 days prior to surgery, ng/mL, median (IQR) | | | |
| Minimum | 0.03 (0.02, 0.03) | 0.01 (0.01, 0.01) | <.001 |
| Maximum | 0.03 (0.03, 0.03) | 0.01 (0.01, 0.03) | <.001 |
| Average | 0.03 (0.03, 0.03) | 0.01 (0.01, 0.02) | <.001 |
| Variance | 0.0 (0.0, 0.0) | 0.0 (0.0, 0.0) | <.001 |
| Count | 0 (0, 0) | 0 (0, 0) | <.001 |
| Serum Troponin T within 8-365 days prior to surgery, ng/mL, median (IQR) | | | |
| Minimum | 0.03 (0.01, 0.03) | 0.01 (0.01, 0.01) | <.001 |
| Maximum | 0.03 (0.03, 0.03) | 0.01 (0.01, 0.02) | <.001 |
| Average | 0.03 (0.03, 0.03) | 0.01 (0.01, 0.01) | <.001 |
| Variance | 0.0 (0.0, 0.0) | 0.0 (0.0, 0.0) | <.001 |
| Count | 0 (0, 0) | 0 (0, 0) | <.001 |
| Reference estimated glomerular filtration rate, median (IQR) | 96.6 (83.8, 108.2) | 99.8 (86.2, 110.8) | <.001 |
| Reference serum creatinine, mg/dL, median (IQR) | 0.8 (0.7, 1.0) | 0.8 (0.7, 1.0) | 0.008 |

Abbreviation: SD, standard deviation; IQR, interquartile range; OB, obstetrician; ACE, Angiotensin-converting enzyme.
a Race and ethnicity were self-reported.
b Other races include American Indian or Alaska Native, Asian, Native Hawaiian or Pacific Islander, and multiracial.

[c] Other surgery type includes plastic surgery, burn surgery, pediatric surgery, transplantation, ophthalmology, medicine gastroenterology, and interventional cardiology.

**eTable 2. Patient characteristics**

| Variables | UFH GNV | | | UFH JAX | | |
|---|---|---|---|---|---|---|
| | Training | Validation | Test | Training | Validation | Test |
| Number of patients, n | 39,582 | 4,397 | 18,848 | 14,846 | 1,649 | 7,068 |
| Number of encounters, n | 51,953 | 5,565 | 22,332 | 18,502 | 2,023 | 8,111 |
| Age in years, mean (SD)[b] | 57 (17) | 57 (17) | 58 (17)[a] | 52 (17) | 53 (17)[a] | 53 (17)[a] |
| Sex, n (%)[b] | | | | | | |
| Male | 19,884 (50) | 2,220 (50) | 9,420 (50) | 7,642 (51) | 863 (52) | 3,671 (52) |
| Female | 19,698 (50) | 2,177 (50) | 9,428 (50) | 7,204 (49) | 786 (48) | 3,397 (48) |
| Race, n (%)[b,c] | | | | | | |
| White | 30,947 (78) | 3,490 (80) | 14,677 (78) | 8,675 (59) | 972 (59) | 4,133 (58) |
| African American | 5,474 (14) | 583 (13) | 2,628 (14) | 5,158 (35) | 560 (34) | 2,407 (34) |
| Other[d] | 2,521 (6) | 264 (6) | 1,216 (6) | 952 (6) | 113 (7) | 484 (7) |
| Missing | 640 (2) | 60 (1) | 327 (2) | 61 (0) | 4 (0) | 44 (1) |
| Ethnicity, n (%)[b,c] | | | | | | |
| Non-Hispanic | 37,165 (94) | 4,078 (93)[a] | 17,421 (93)[a] | 14,079 (95) | 1,559 (95) | 6,639 (94)[a] |
| Hispanic | 1,707 (4) | 254 (6)[a] | 980 (5)[a] | 676 (5) | 85 (5) | 376 (5)[a] |
| Missing | 710 (2) | 65 (1) | 447 (2)[a] | 91 (0) | 5 (0) | 53 (1) |
| Marital Status, n (%)[b] | | | | | | |
| Married | 19,182 (48) | 2,001 (45)[a] | 8,815 (47)[a] | 5,055 (34) | 566 (34) | 2,577 (37)[a] |
| Single | 11,769 (30) | 1,256 (29) | 5,333 (28)[a] | 5,392 (36) | 591 (36) | 2,492 (35) |
| Divorced | 6,084 (15) | 596 (13)[a] | 2,630 (14)[a] | 3,875 (26) | 437 (26) | 1,776 (25) |
| Missing | 2,547 (7) | 544 (13)[a] | 2,070 (11)[a] | 524 (4) | 55 (4) | 223 (3) |
| Insurance, n (%)[b] | | | | | | |
| Medicare | 17,767 (45) | 1,987 (45) | 8,841 (47)[a] | 4,408 (30) | 517 (31) | 2,208 (31) |
| Private | 12,311 (31) | 1,288 (29)[a] | 5,497 (29)[a] | 4,308 (29) | 492 (30) | 2,181 (31)[a] |
| Medicaid | 6,296 (16) | 731 (17) | 2,894 (15) | 5,908 (40) | 598 (36)[a] | 2,499 (35)[a] |
| Uninsured | 3,208 (8) | 391 (9) | 1616 (9) | 222 (1) | 42 (3)[a] | 180 (3)[a] |
| Complications, n (%)[e] | | | | | | |

| Variables | UFH GNV | | | UFH JAX | | |
|---|---|---|---|---|---|---|
| | Training | Validation | Test | Training | Validation | Test |
| Acute kidney injury | 7,924 (15) | 932 (17)[a] | 3,917 (18)[a] | 2,577 (14) | 266 (13) | 1,048 (13) |
| Cardiovascular complications | 6,419 (12) | 878 (16)[a] | 3,911 (18)[a] | 2,104 (11) | 234 (12) | 788 (10)[a] |
| Neurological complications, including delirium | 8,873 (17) | 1,178 (21)[a] | 5,135 (23)[a] | 2,199 (12) | 240 (12) | 973 (12) |
| Prolonged ICU stay | 15,049 (29) | 1,610 (29) | 7,657 (34)[a] | 4,372 (24) | 512 (25) | 2,013 (25) |
| Prolonged mechanical ventilation | 4,667 (9) | 461 (8) | 1,833 (8)[a] | 1,492 (8) | 135 (7) | 489 (6)[a] |
| Sepsis | 3,958 (8) | 542 (10)[a] | 2,195 (10)[a] | 1,601 (9) | 202 (10) | 669 (8) |
| Venous thromboembolism | 2,483 (5) | 310 (6)[a] | 1,423 (6)[a] | 721 (4) | 69 (3) | 255 (3)[a] |
| Wound complications | 8,088 (16) | 1,174 (21)[a] | 5,037 (23)[a] | 2,594 (14) | 290 (14) | 1,107 (14) |
| Hospital mortality | 952 (2) | 93 (2) | 383 (2) | 329 (2) | 27 (1) | 97 (1)[a] |

Abbreviation: ICU, intensive care unit.
[a] p<=0.05 comparing validation to training data for each site.
[b] Data were reported based on values calculated at the latest hospital admission.
[c] Race and ethnicity were self-reported.
[d] Other races include American Indian or Alaska Native, Asian, Native Hawaiian or Pacific Islander, and multiracial.
[e] Data were reported based on postoperative complication status for each surgical procedure.

**eTable 3: Comparison of AUPRC for central learning and federated learning models**

| | | UFH GNV | | | | UFH JAX | | | |
|---|---|---|---|---|---|---|---|---|---|
| Outcome | Period | CL | FedAvg | FedProx | SCAFFOLD | CL | FedAvg | FedProx | SCAFFOLD |
| Prolonged ICU stay | PreOp | 0.82 (0.81-0.83) | 0.84 (0.83-0.84) | 0.84 (0.83-0.85) | 0.84 (0.83-0.84) | 0.77 (0.75-0.78) | 0.72 (0.70-0.73) | 0.71 (0.69-0.72) | 0.71 (0.70-0.73) |
| | PostOp | 0.86 (0.85-0.86) | 0.87 (0.86-0.87) | 0.87 (0.86-0.87) | 0.87 (0.87-0.88) | 0.80 (0.78-0.81) | 0.74 (0.72-0.75) | 0.73 (0.71-0.75) | 0.76 (0.74-0.78) |
| Sepsis | PreOp | 0.53 (0.51-0.55) | 0.53 (0.51-0.55) | 0.54 (0.52-0.56) | 0.54 (0.52-0.56) | 0.52 (0.48-0.55) | 0.49 (0.45-0.52) | 0.49 (0.45-0.52) | 0.50 (0.46-0.53) |
| | PostOp | 0.55 (0.53-0.571) | 0.56 (0.54-0.58) | 0.56 (0.54-0.58) | 0.56 (0.54-0.58) | 0.53 (0.49-0.56) | 0.52 (0.48-0.55) | 0.51 (0.48-0.55) | 0.51 (0.48-0.55) |
| Cardiovascular complication | PreOp | 0.52 (0.51-0.54) | 0.52 (0.51-0.54) | 0.53 (0.52-0.54) | 0.53 (0.51-0.54) | 0.36 (0.33-0.39) | 0.32 (0.30-0.35) | 0.33 (0.30-0.36) | 0.34 (0.32-0.37) |
| | PostOp | 0.59 (0.57-0.60) | 0.59 (0.58-0.61) | 0.59 (0.58-0.61) | 0.59 (0.58-0.60) | 0.43 (0.40-0.46) | 0.43 (0.40-0.46) | 0.42 (0.39-0.45) | 0.43 (0.40-0.46) |
| Venous thromboembolism | PreOp | 0.26 (0.24-0.27) | 0.26 (0.25-0.28) | 0.26 (0.24-0.28) | 0.27 (0.25-0.29) | 0.14 (0.12-0.18) | 0.12 (0.10-0.15) | 0.13 (0.10-0.16) | 0.14 (0.11-0.18) |
| | PostOp | 0.25 (0.23-0.27) | 0.28 (0.26-0.30) | 0.27 (0.25-0.29) | 0.29 (0.27-0.31) | 0.16 (0.13-0.20) | 0.15 (0.12-0.18) | 0.14 (0.11-0.17) | 0.14 (0.12-0.18) |
| Prolonged mechanical ventilation | PreOp | 0.56 (0.54-0.58) | 0.56 (0.54-0.58) | 0.56 (0.54-0.58) | 0.56 (0.54-0.58) | 0.43 (0.39-0.47) | 0.39 (0.35-0.43) | 0.38 (0.34-0.42) | 0.40 (0.36-0.44) |
| | PostOp | 0.60 (0.58-0.62) | 0.62 (0.61-0.64) | 0.62 (0.60-0.64) | 0.62 (0.60-0.63) | 0.50 (0.46-0.53) | 0.46 (0.43-0.50) | 0.45 (0.41-0.49) | 0.49 (0.45-0.53) |
| Neurological complications, including delirium | PreOp | 0.67 (0.66-0.68) | 0.67 (0.66-0.68) | 0.68 (0.67-0.69) | 0.68 (0.67-0.69) | 0.48 (0.45-0.51) | 0.47 (0.44-0.50) | 0.47 (0.44-0.50) | 0.48 (0.46-0.51) |
| | PostOp | 0.67 (0.66-0.68) | 0.68 (0.67-0.70) | 0.68 (0.67-0.69) | 0.69 (0.68-0.70) | 0.49 (0.46-0.52) | 0.48 (0.46-0.51) | 0.48 (0.45-0.51) | 0.49 (0.46-0.52) |
| Wound complications | PreOp | 0.58 (0.57-0.60) | 0.58 (0.57-0.59) | 0.59 (0.58-0.61) | 0.60 (0.58-0.61) | 0.37 (0.35-0.40) | 0.36 (0.34-0.39) | 0.35 (0.33-0.38) | 0.37 (0.34-0.39) |
| | PostOp | 0.59 (0.58-0.60) | 0.60 (0.59-0.61) | 0.61 (0.59-0.62) | 0.61 (0.60-0.62) | 0.38 (0.36-0.41) | 0.37 (0.35-0.39) | 0.37 (0.35-0.40) | 0.39 (0.37-0.42) |
| Acute kidney injury | PreOp | 0.52 (0.51-0.53) | 0.53 (0.51-0.54) | 0.53 (0.51-0.54) | 0.53 (0.51-0.54) | 0.40 (0.37-0.43) | 0.41 (0.38-0.44) | 0.40 (0.38-0.43) | 0.41 (0.38-0.43) |
| | PostOp | 0.53 (0.52-0.54) | 0.55 (0.53-0.56) | 0.54 (0.53-0.56) | 0.56 (0.54-0.57) | 0.42 (0.40-0.45) | 0.43 (0.41-0.46) | 0.43 (0.40-0.45) | 0.44 (0.41-0.46) |
| Hospital mortality | PreOp | 0.18 (0.15-0.21) | 0.17 (0.15-0.21) | 0.18 (0.15-0.21) | 0.17 (0.14-0.19) | 0.16 (0.12-0.22) | 0.14 (0.10-0.20) | 0.14 (0.10-0.19) | 0.14 (0.10-0.19) |
| | PostOp | 0.19 (0.16-0.22) | 0.22 (0.18-0.25) | 0.20 (0.17-0.23) | 0.20 (0.17-0.24) | 0.24 (0.17-0.32) | 0.22 (0.16-0.30) | 0.21 (0.16-0.29) | 0.22 (0.17-0.30) |

Abbreviation: ICU, intensive care unit; CL, central learning.

**eTable 4: Comparison of AUPRC for local learning and federated learning models.**

|  |  | Preoperative models | | Postoperative models | |
|---|---|---|---|---|---|
| **Outcome** | **Model** | **GNV test data** | **JAX test data** | **GNV test data** | **JAX test data** |
| Prolonged ICU stay | GNV Model | 0.83 (0.83-0.84) | 0.62 (0.60-0.63) | 0.87 (0.87-0.88) | 0.71 (0.69-0.73) |
|  | JAX Model | 0.58 (0.57-0.59) | 0.76 (0.74-0.78) | 0.80 (0.79-0.80) | 0.80 (0.78-0.81) |
|  | SCAFFOLD | 0.84 (0.83-0.84) | 0.71 (0.70-0.73) | 0.87 (0.87-0.88) | 0.76 (0.74-0.78) |
| Sepsis | GNV Model | 0.53 (0.51-0.55) | 0.47 (0.44-0.51) | 0.56 (0.54-0.57) | 0.51 (0.48-0.54) |
|  | JAX Model | 0.42 (0.40-0.44) | 0.52 (0.49-0.55) | 0.44 (0.42-0.46) | 0.52 (0.49-0.56) |
|  | SCAFFOLD | 0.54 (0.52-0.56) | 0.50 (0.46-0.53) | 0.56 (0.54-0.58) | 0.51 (0.48-0.55) |
| Cardiovascular complication | GNV Model | 0.52 (0.50-0.53) | 0.28 (0.26-0.31) | 0.58 (0.57-0.60) | 0.41 (0.38-0.44) |
|  | JAX Model | 0.34 (0.33-0.36) | 0.36 (0.34-0.39) | 0.54 (0.53-0.60) | 0.43 (0.40-0.46) |
|  | SCAFFOLD | 0.53 (0.51-0.54) | 0.34 (0.32-0.37) | 0.59 (0.58-0.60) | 0.43 (0.40-0.46) |
| Venous thromboembolism | GNV Model | 0.25 (0.23-0.27) | 0.12 (0.10-0.15) | 0.27 (0.25-0.29) | 0.14 (0.11-0.17) |
|  | JAX Model | 0.16 (0.15-0.18) | 0.13 (0.11-0.16) | 0.19 (0.18-0.21) | 0.14 (0.12-0.17) |
|  | SCAFFOLD | 0.27 (0.25-0.29) | 0.14 (0.11-0.18) | 0.29 (0.27-0.31) | 0.14 (0.12-0.18) |
| Prolonged mechanical ventilation | GNV Model | 0.53 (0.51-0.55) | 0.30 (0.27-0.34) | 0.61 (0.59-0.62) | 0.41 (0.38-0.45) |
|  | JAX Model | 0.40 (0.38-0.42) | 0.41 (0.37-0.45) | 0.49 (0.47-0.51) | 0.51 (0.47-0.55) |
|  | SCAFFOLD | 0.56 (0.54-0.58) | 0.40 (0.36-0.44) | 0.62 (0.60-0.63) | 0.49 (0.45-0.53) |
| Neurological complications, including delirium | GNV Model | 0.66 (0.65-0.68) | 0.43 (0.40-0.46) | 0.68 (0.67-0.69) | 0.46 (0.43-0.48) |
|  | JAX Model | 0.47 (0.46-0.49) | 0.43 (0.41-0.46) | 0.52 (0.51-0.53) | 0.46 (0.43-0.48) |
|  | SCAFFOLD | 0.68 (0.67-0.69) | 0.48 (0.46-0.51) | 0.69 (0.68-0.70) | 0.49 (0.46-0.52) |
| Wound complications | GNV Model | 0.58 (0.57-0.59) | 0.32 (0.30-0.34) | 0.60 (0.59-0.61) | 0.34 (0.32-0.37) |
|  | JAX Model | 0.38 (0.37-0.39) | 0.31 (0.29-0.33) | 0.42 (0.41-0.43) | 0.32 (0.30-0.35) |
|  | SCAFFOLD | 0.60 (0.58-0.61) | 0.37 (0.34-0.39) | 0.61 (0.60-0.62) | 0.39 (0.37-0.42) |
| Acute kidney injury | GNV Model | 0.52 (0.51-0.54) | 0.40 (0.37-0.42) | 0.55 (0.53-0.56) | 0.43 (0.40-0.45) |
|  | JAX Model | 0.33 (0.32-0.34) | 0.38 (0.36-0.41) | 0.45 (0.44-0.46) | 0.42 (0.39-0.44) |
|  | SCAFFOLD | 0.53 (0.51-0.54) | 0.41 (0.38-0.43) | 0.55 (0.54-0.57) | 0.44 (0.41-0.46) |
| Hospital mortality | GNV Model | 0.16 (0.14-0.19) | 0.10 (0.07-0.15) | 0.20 (0.17-0.23) | 0.19 (0.14-0.26) |
|  | JAX Model | 0.13 | 0.14 | 0.16 | 0.20 |

| Outcome | Model | Preoperative models | | Postoperative models | |
|---|---|---|---|---|---|
| | | GNV test data | JAX test data | GNV test data | JAX test data |
| | | (0.12-0.16) | (0.10-0.20) | (0.13-0.19) | (0.15-0.28) |
| | SCAFFOLD | 0.17 | 0.14 | 0.20 | 0.22 |
| | | (0.14-0.19) | (0.10-0.19) | (0.17-0.24) | (0.17-0.30) |

Abbreviation: ICU, intensive care unit.

**eTable 5: Subgroup analysis of AUROC with 95% confidence interval for federated learning model (SCAFFOLD) based on sex at each center**

| | | UFH GNV | | UFH JAX | |
|---|---|---|---|---|---|
| **Outcome** | **Period** | **Female** | **Male** | **Female** | **Male** |
| Prolonged ICU stay | PreOp | 0.90 (0.90-0.91) | 0.90 (0.89-0.90) | 0.86 (0.85-0.87) | 0.86 (0.85-0.87) |
| | PostOp | 0.92 (0.92-0.93) | 0.92 (0.91-0.92) | 0.89 (0.88-0.90) | 0.90 (0.89-0.90) |
| Sepsis | PreOp | 0.89 (0.88-0.90) | 0.87 (0.86-0.88)[a] | 0.90 (0.88-0.92) | 0.87 (0.85-0.88)[a] |
| | PostOp | 0.90 (0.89-0.91) | 0.88 (0.87-0.88)[a] | 0.91 (0.90-0.93) | 0.87 (0.86-0.89)[a] |
| Cardiovascular complication | PreOp | 0.81 (0.80-0.82) | 0.82 (0.82-0.83)[a] | 0.76 (0.74-0.78) | 0.81 (0.80-0.83)[a] |
| | PostOp | 0.84 (0.84-0.85) | 0.85 (0.85-0.86) | 0.82 (0.80-0.84) | 0.86 (0.84-0.87)[a] |
| Venous thromboembolism | PreOp | 0.85 (0.84-0.87) | 0.81 (0.80-0.82)[a] | 0.83 (0.79-0.86) | 0.75 (0.72-0.78)[a] |
| | PostOp | 0.86 (0.85-0.87) | 0.82 (0.81-0.82)[a] | 0.85 (0.81-0.88) | 0.80 (0.78-0.83) |
| Prolonged mechanical ventilation | PreOp | 0.90 (0.90-0.91) | 0.90 (0.89-0.91) | 0.84 (0.81-0.87) | 0.83 (0.81-0.85) |
| | PostOp | 0.92 (0.91-0.93) | 0.92 (0.91-0.92) | 0.86 (0.83-0.89) | 0.86 (0.84-0.88) |
| Neurological complications, including delirium | PreOp | 0.86 (0.85-0.86) | 0.85 (0.84-0.86) | 0.85 (0.83-0.87) | 0.83 (0.81-0.84) |
| | PostOp | 0.86 (0.85-0.87) | 0.86 (0.85-0.86) | 0.86 (0.84-0.87) | 0.83 (0.82-0.84) |
| Wound complications | PreOp | 0.80 (0.79-0.81) | 0.79 (0.78-0.80) | 0.70 (0.68-0.73) | 0.71 (0.70-0.73) |
| | PostOp | 0.81 (0.80-0.82) | 0.80 (0.79-0.81) | 0.73 (0.71-0.75) | 0.74 (0.72-0.76) |
| Acute kidney injury | PreOp | 0.83 (0.82-0.84) | 0.81 (0.81-0.82) | 0.81 (0.79-0.83) | 0.78 (0.76-0.79)[a] |
| | PostOp | 0.84 (0.83-0.85) | 0.83 (0.82-0.83) | 0.82 (0.81-0.84) | 0.80 (0.78-0.82) |
| Hospital mortality | PreOp | 0.89 (0.87-0.91) | 0.88 (0.87-0.90) | 0.94 (0.91-0.97) | 0.87 (0.84-0.90)[a] |
| | PostOp | 0.91 (0.89-0.92) | 0.91 (0.89-0.92) | 0.98 (0.96-0.99) | 0.89 (0.85-0.92)[a] |

[a] p<=0.05 comparing female to male patients for each site.
Abbreviation: ICU, intensive care unit; PreOp, preoperative; PostOp, postoperative.

**eTable 6: Subgroup analysis of AUROC with 95% confidence interval for federated learning model (SCAFFOLD) based on race at each center**

|  |  | UFH GNV | | UFH JAX | |
|---|---|---|---|---|---|
| **Outcome** | **Period** | **African American** | **Non-African American** | **African American** | **Non-African American** |
| Prolonged ICU stay | PreOp | 0.90 (0.89-0.91) | 0.90 (0.90-0.91) | 0.87 (0.86-0.88) | 0.87 (0.86-0.87) |
|  | PostOp | 0.92 (0.91-0.92) | 0.92 (0.92-0.92) | 0.90 (0.89-0.91) | 0.89 (0.89-0.90) |
| Sepsis | PreOp | 0.90 (0.88-0.91) | 0.874 (0.87-0.88)[a] | 0.89 (0.87-0.91) | 0.88 (0.86-0.89) |
|  | PostOp | 0.91 (0.90-0.92) | 0.88 (0.88-0.89)[a] | 0.90 (0.88-0.92) | 0.88 (0.87-0.90) |
| Cardiovascular complication | PreOp | 0.82 (0.81-0.84) | 0.82 (0.81-0.82) | 0.78 (0.76-0.81) | 0.79 (0.77-0.81) |
|  | PostOp | 0.85 (0.84-0.87) | 0.85 (0.84-0.86) | 0.84 (0.82-0.86) | 0.84 (0.83-0.86) |
| Venous thromboembolism | PreOp | 0.85 (0.83-0.87) | 0.83 (0.82-0.84) | 0.77 (0.73-0.80) | 0.80 (0.77-0.83) |
|  | PostOp | 0.85 (0.82-0.87) | 0.84 (0.83-0.85) | 0.83 (0.79-0.86) | 0.82 (0.80-0.85) |
| Prolonged mechanical ventilation | PreOp | 0.90 (0.88-0.92) | 0.90 (0.90-0.91) | 0.84 (0.81-0.87) | 0.83 (0.81-0.85) |
|  | PostOp | 0.91 (0.89-0.92) | 0.92 (0.91-0.92) | 0.86 (0.83-0.88) | 0.87 (0.85-0.88) |
| Neurological complications, including delirium | PreOp | 0.87 (0.86-0.88) | 0.85 (0.85-0.86) | 0.85 (0.84-0.87) | 0.83 (0.82-0.85) |
|  | PostOp | 0.87 (0.86-0.88) | 0.86 (0.85-0.86) | 0.86 (0.84-0.88) | 0.83 (0.82-0.85) |
| Wound complications | PreOp | 0.78 (0.77-0.80) | 0.80 (0.79-0.81) | 0.69 (0.67-0.72) | 0.72 (0.70-0.74) |
|  | PostOp | 0.80 (0.78-0.81) | 0.81 (0.80-0.82) | 0.74 (0.71-0.76) | 0.74 (0.72-0.75) |
| Acute kidney injury | PreOp | 0.81 (0.80-0.83) | 0.82 (0.82-0.83) | 0.81 (0.78-0.82) | 0.78 (0.76-0.79) |
|  | PostOp | 0.83 (0.81-0.84) | 0.83 (0.83-0.84) | 0.83 (0.81-0.84) | 0.80 (0.78-0.81) |
| Hospital mortality | PreOp | 0.92 (0.89-0.94) | 0.88 (0.87-0.90)[a] | 0.94 (0.91-0.96) | 0.88 (0.85-0.91)[a] |
|  | PostOp | 0.93 (0.91-0.95) | 0.90 (0.89-0.91) | 0.97 (0.96-0.98) | 0.89 (0.86-0.92)[a] |

[a] p<=0.05 comparing African American to Non-African American patients for each site.
Abbreviation: ICU, intensive care unit; PreOp, preoperative; PostOp, postoperative.

**eTable 7: Subgroup analysis of AUROC with 95% confidence interval for federated learning model (SCAFFOLD) based on age at each center**

|  |  | UFH GNV | | UFH JAX | |
|---|---|---|---|---|---|
| **Outcome** | **Period** | **Age <= 65** | **Age > 65** | **Age <= 65** | **Age > 65** |
| Prolonged ICU stay | PreOp | 0.91 (0.90-0.91) | 0.89 (0.89-0.90)[a] | 0.87 (0.86-0.88) | 0.85 (0.84-0.87) |
|  | PostOp | 0.92 (0.92-0.93) | 0.91 (0.91-0.92)[a] | 0.90 (0.89-0.91) | 0.88 (0.86-0.89) |
| Sepsis | PreOp | 0.884 (0.876-0.892) | 0.87 (0.86-0.88)[a] | 0.88 (0.87-0.89) | 0.89 (0.87-0.91) |
|  | PostOp | 0.89 (0.88-0.90) | 0.88 (0.87-0.89) | 0.89 (0.88-0.90) | 0.90 (0.88-0.92) |
| Cardiovascular complication | PreOp | 0.83 (0.82-0.84) | 0.79 (0.78-0.80)[a] | 0.80 (0.78-0.81) | 0.75 (0.72-0.78) |
|  | PostOp | 0.86 (0.85-0.87) | 0.83 (0.82-0.84)[a] | 0.85 (0.83-0.86) | 0.80 (0.78-0.83) |
| Venous thromboembolism | PreOp | 0.85 (0.84-0.86) | 0.80 (0.79-0.82) | 0.80 (0.77-0.82) | 0.76 (0.72-0.80) |
|  | PostOp | 0.85 (0.84-0.86) | 0.82 (0.80-0.83) | 0.83 (0.80-0.85) | 0.82 (0.78-0.85) |
| Prolonged mechanical ventilation | PreOp | 0.91 (0.91-0.92) | 0.89 (0.87-0.90) | 0.84 (0.82-0.86) | 0.83 (0.80-0.86) |
|  | PostOp | 0.93 (0.92-0.93) | 0.90 (0.89-0.91) | 0.86 (0.84-0.88) | 0.86 (0.83-0.90) |
| Neurological complications, including delirium | PreOp | 0.87 (0.86-0.88) | 0.83 (0.82-0.83) | 0.86 (0.84-0.87) | 0.78 (0.75-0.80) |
|  | PostOp | 0.87 (0.87-0.88) | 0.83 (0.82-0.84) | 0.86 (0.84-0.87) | 0.79 (0.76-0.81) |
| Wound complications | PreOp | 0.79 (0.78-0.80) | 0.81 (0.80-0.82) | 0.71 (0.70-0.73) | 0.70 (0.67-0.73) |
|  | PostOp | 0.80 (0.79-0.81) | 0.82 (0.81-0.83) | 0.74 (0.72-0.75) | 0.73 (0.70-0.76) |
| Acute kidney injury | PreOp | 0.83 (0.83-0.84) | 0.80 (0.79-0.81) | 0.79 (0.78-0.81) | 0.76 (0.73-0.78) |
|  | PostOp | 0.84 (0.84-0.85) | 0.81 (0.80-0.82) | 0.82 (0.80-0.83) | 0.77 (0.75-0.79) |
| Hospital mortality | PreOp | 0.92 (0.90-0.93) | 0.84 (0.82-0.87) | 0.91 (0.88-0.94) | 0.84 (0.80-0.88) |
|  | PostOp | 0.93 (0.91-0.94) | 0.88 (0.86-0.89) | 0.93 (0.90-0.96) | 0.89 (0.84-0.92) |

[a] $p \leq 0.05$ comparing African American to Non-African American patients for each site.
Abbreviation: ICU, intensive care unit; PreOp, preoperative; PostOp, postoperative.

**eTable 8: Comparison of AUROC for federated learning preoperative models with different training sample sizes**

| Outcome | UFH GNV | | UFH JAX | |
|---|---|---|---|---|
| | SCAFFOLD model trained using raw data size | SCAFFOLD model trained using equal size, mean (SD) | SCAFFOLD model trained using raw data size | SCAFFOLD model trained using equal size, mean (SD) |
| Prolonged ICU stay | 0.90 | 0.89 (0.002) | 0.87 | 0.89 (0.003) |
| Sepsis | 0.88 | 0.87 (0.002) | 0.88 | 0.88 (0.002) |
| Cardiovascular complication | 0.82 | 0.81 (0.002) | 0.79 | 0.79 (0.002) |
| Venous thromboembolism | 0.83 | 0.81 (0.004) | 0.79 | 0.80 (0.005) |
| Prolonged mechanical ventilation | 0.90 | 0.89 (0.005) | 0.84 | 0.84 (0.004) |
| Neurological complications, including delirium | 0.85 | 0.84 (0.002) | 0.84 | 0.84 (0.003) |
| Wound complications | 0.80 | 0.78 (0.003) | 0.71 | 0.71 (0.002) |
| Acute kidney injury | 0.82 | 0.81 (0.001) | 0.79 | 0.79 (0.003) |
| Hospital mortality | 0.89 | 0.88 (0.011) | 0.90 | 0.91 (0.002) |

Abbreviation: ICU, intensive care unit; SD, standard deviation.